\documentclass[10pt,twocolumn,letterpaper]{article}

\usepackage{iccv}
\usepackage{times}
\usepackage{epsfig}
\usepackage{graphicx}
\usepackage{amsmath}
\usepackage{amssymb}
\usepackage{capt-of}
\usepackage{multirow}
\usepackage{amsthm}
\usepackage{comment}
\usepackage{diagbox}
\usepackage[accsupp]{axessibility}
\makeatletter
\@namedef{ver@everyshi.sty}{}
\makeatother
\usepackage{pgfplots}
\pgfplotsset{width=10cm,compat=1.9}

\newtheorem*{definition}{Definition}

\newcommand\blfootnote[1]{%
  \begingroup
  \renewcommand\thefootnote{}\footnote{#1}%
  \addtocounter{footnote}{-1}%
  \endgroup
}

\usepackage[pagebackref=true,breaklinks=true,letterpaper=true,colorlinks,bookmarks=false]{hyperref}

\iccvfinalcopy %

\def\iccvPaperID{8434} %

\usepackage{scalerel,stackengine}
\stackMath
\newcommand\reallywidehat[1]{%
\savestack{\tmpbox}{\stretchto{%
  \scaleto{%
    \scalerel*[\widthof{\ensuremath{#1}}]{\kern-.6pt\bigwedge\kern-.6pt}%
    {\rule[-\textheight/2]{1ex}{\textheight}}%
  }{\textheight}%
}{0.5ex}}%
\stackon[1pt]{#1}{\tmpbox}%
}

\begin{document}

\title{CAPTRA: CAtegory-level Pose Tracking \\for Rigid and Articulated Objects from Point Clouds}

\author{
Yijia Weng\textsuperscript{1*} \quad 
He Wang\textsuperscript{1,2,5*$^\dagger$} \quad 
Qiang Zhou \textsuperscript{4} \quad 
Yuzhe Qin\textsuperscript{3} \quad 
Yueqi Duan\textsuperscript{2}\\
Qingnan Fan\textsuperscript{2,6} \quad 
Baoquan Chen\textsuperscript{1} \quad 
Hao Su\textsuperscript{3} \quad 
Leonidas J.~Guibas\textsuperscript{2}\\
\textsuperscript{1}CFCS, Peking University \quad 
\textsuperscript{2}Stanford University  \quad
\textsuperscript{3}UCSD   \\
\textsuperscript{4}Shandong University \quad
\textsuperscript{5}Beijing Institute for General AI \quad \textsuperscript{6}Tencent AI Lab
\\}

\twocolumn[{%
\renewcommand\twocolumn[1][]{#1}
\maketitle

\ificcvfinal\thispagestyle{empty}\fi

\vspace{-3em}
\begin{center}
    \centering
    \includegraphics[width=0.8\linewidth]{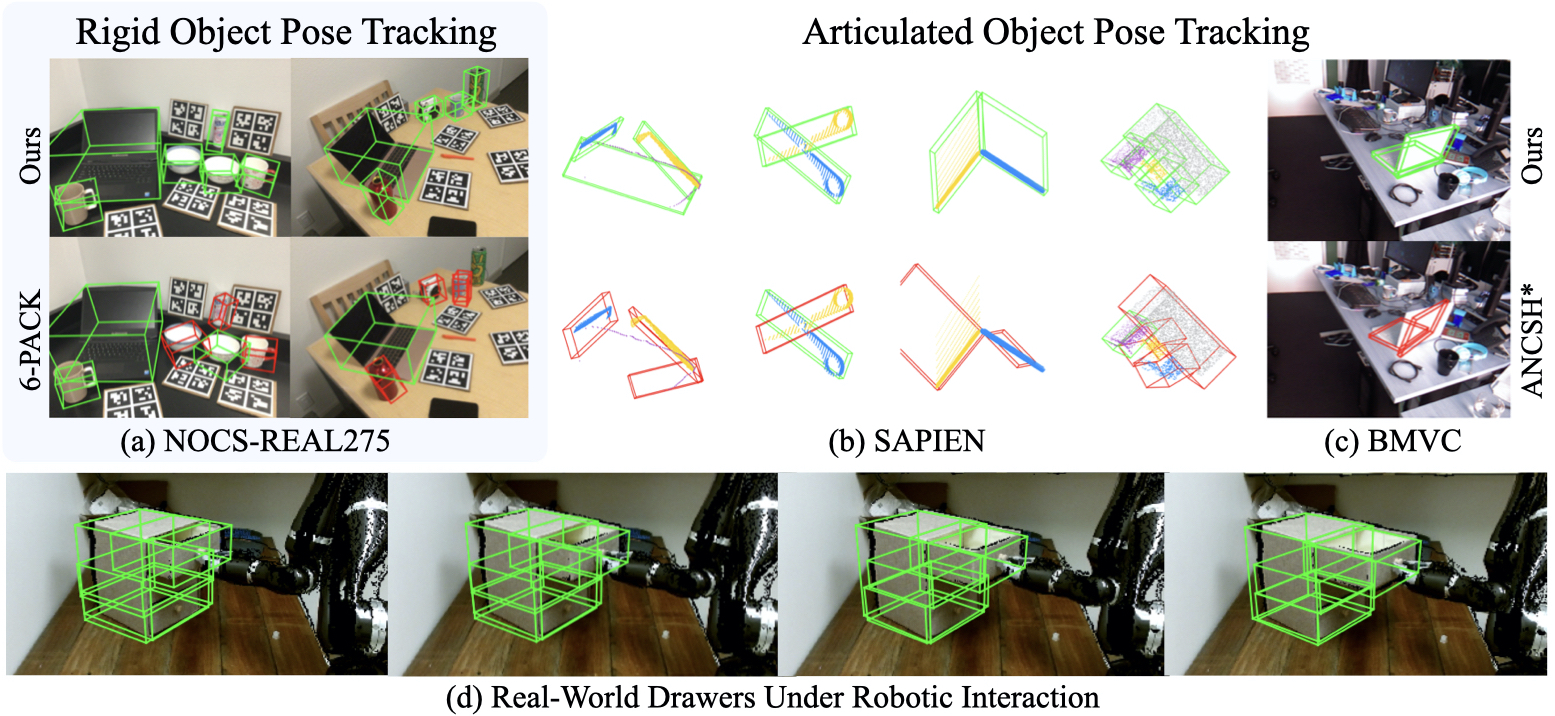}
    \captionof{figure}
    {Our method tracks 9DoF category-level poses (3D rotation, 3D translation, and 3D size) of novel rigid objects as well as parts in articulated objects from live point cloud streams. We demonstrate: 
    (a) our method can reliably track rigid object poses from the challenging NOCS-REAL275 dataset \cite{wang2019normalized}; 
    (b) our method can perfectly track articulated objects with big global and articulated motions from the SAPIEN datasets\cite{xiang2020sapien};
    (c)(d) trained only on SAPIEN, our model can directly generalize to novel real laptops from BMVC dataset\cite{BMVC2015_181}, and novel real drawers under robotic interaction. In all cases, our method significantly outperforms the previous state-of-the-arts and baselines.
    Here we visualize the estimated 9DoF poses as 3D bounding boxes: green boxes indicate in tracking whereas red boxes indicate off tracking.}
    \label{fig:teaser}
\end{center}
}]

\blfootnote{*: equal contributions, $\dagger$: corresponding author}
\blfootnote{Project page:~\href{https://yijiaweng.github.io/CAPTRA}{https://yijiaweng.github.io/CAPTRA}}

\begin{abstract}
In this work, we tackle the problem of category-level online pose tracking of objects from point cloud sequences. For the first time,  we propose a unified framework that can handle 9DoF pose tracking for novel rigid object instances as well as per-part pose tracking for articulated objects from known categories. Here the 9DoF pose, comprising 6D pose and 3D size, is equivalent to a 3D amodal bounding box representation with free 6D pose. Given the depth point cloud at the current frame and the estimated pose from the last frame, our novel end-to-end pipeline learns to accurately update the pose. Our pipeline is composed of three modules: 1) a pose canonicalization module that normalizes the pose of the input depth point cloud; 2) \emph{RotationNet}, a module that directly regresses small interframe delta rotations; and 3) \emph{CoordinateNet}, a module that predicts the normalized coordinates and segmentation, enabling analytical computation of the 3D size and translation. Leveraging the small pose regime in the pose-canonicalized point clouds, our method integrates the best of both worlds by combining dense coordinate prediction and direct rotation regression, thus yielding an end-to-end differentiable pipeline optimized for 9DoF pose accuracy (without using non-differentiable RANSAC). Our extensive experiments demonstrate that our method achieves new state-of-the-art performance on category-level rigid object pose (NOCS-REAL275 \cite{wang2019normalized}) and articulated object pose benchmarks (SAPIEN \cite{xiang2020sapien}, BMVC \cite{BMVC2015_181}) at the fastest FPS $\sim 12$. 
\end{abstract}
\vspace{-2mm}

\vspace{-3mm}
\section{Introduction}
\vspace{-2mm}

Object pose estimation is crucial for a variety of
computer vision and robotics applications, such as 3D scene understanding, robotic manipulation and augmented reality. 
The majority of object pose estimation works, \textit{e.g.}, \cite{xiang2017posecnn, rad2017bb8}, mainly lie in instance-level estimation, where the task is to estimate poses for objects from a small set of a priori known instances, thus preventing them from perceiving the poses of the vast diversity of objects in our daily life. To mitigate this limitation, Wang \emph{et al.} \cite{wang2019normalized} proposed to generalize the instance-level 6DoF (Degree of Freedom) object pose estimation problem to a category-level 9DoF pose estimation problem that takes into account the traditional 6D object pose (rotation, translation) as well as 3D object size. The proposed method in \cite{wang2019normalized} can handle novel object instances in known categories without  requiring CAD models of the objects. Going beyond rigid objects and in the same spirit,  Li \emph{et al.} \cite{li2020categorylevel} proposed to estimate category-level per-part 9DoF poses for articulated objects, such as laptops, drawers and eyeglasses. %

While most of the existing category-level pose estimation works focus on single-frame estimation, we believe that temporally smooth pose tracking is more useful for many robotics applications, \textit{e.g.}, instant feedback control, as well as AR applications. In this work, we tackle a problem named CAPTRA --- \textit{CAtegory-level Pose Tracking for Rigid and Articulated Objects}, from a live point cloud stream. Given an initial object pose at the first frame, our task is to continuously track the 9DoF pose for rigid objects or each individual rigid part of an articulated object. The most related work to ours is 6-PACK \cite{wang20196}, which tackles the problem of category-level 6D pose tracking only for rigid objects (see the related work section for detailed comparisons).

To accurately track 9DoF poses, we consider two types of approaches: coordinate-based approaches widely used in object pose \cite{brachmann2017dsac, wang2019normalized, li2020categorylevel} and camera pose estimation \cite{cavallari2017fly} and direct pose regression as in \cite{xiang2017posecnn, wen2020se}.
These two approaches both have pros and cons. 
Coordinate-based methods, which predict dense object coordinates followed by a RANSAC-based pose fitting, are generally more accurate and robust, especially on rotation estimation \cite{umeyama1991least}, benefiting from outlier removal in RANSAC. However, RANSAC-based pose fitting is non-differentiable and time-consuming, which often leads to a bottleneck in its running speed. In contrast, direct pose regression performs an end-to-end pose prediction, thus can achieve very high running speed, at the cost of being more error-prone. Notably, regression of arbitrary rotations $\in SO(3)$ is particularly challenging because of the nonlinearity of the output space.

In this work, we seek to take the best of both worlds and build \textbf{an end-to-end differentiable pipeline for accurate and fast pose tracking}. 
To enable highly accurate pose estimation, we propose to jointly canonicalize the input and output spaces of this estimation problem by transforming the point clouds using the inverse poses from the previous frame.
The poses of the regarding objects/parts in the produced \textbf{pose-canonicalized point clouds} are nearly identical, containing only small residual rotations that span an almost linear space due to the Lie group properties of SO(3) and therefore are regression-friendly.
We thus propose \textbf{RotationNet}, a PointNet++~\cite{qi2017pointnet++} based neural network, to directly regress the small residual rotations.
However, we found scale and translation regression still challenging, partially due to the ambiguity between occlusion and center translation in the partial depth observations,
We instead propose to build \textbf{CoordinateNet} to predict dense normalized coordinates, which contain more accurate information about translation and object size due to their awareness of the category-level shape prior. 
Combining the outputs from both networks, we can analytically compute sizes and translations, yielding an end-to-end differentiable pipeline optimized for 9DoF pose accuracy without using non-differentiable RANSAC. 

By harnessing both approaches, our proposed method gains significant performance improvement on the category-level rigid object pose benchmark and articulated object pose benchmarks. 
On the NOCS-REAL275 dataset \cite{wang2019normalized}, we outperform 6-PACK \cite{wang20196}, the previous state-of-the-art, by 40.03\% absolute improvement on the mean accuracy of $5^\circ 5$cm and 10.52\% absolute improvement on the mean IoU metric.
On the SAPIEN articulated object dataset \cite{xiang2020sapien}, we are the first to perform tracking and outperform the single-frame articulated pose estimation baseline, constructed using ANCSH \cite{li2020categorylevel} and ground truth segmentation masks, by a large margin, \textit{e.g.}, around 20 points on mean accuracy $5^\circ 5$cm in the challenging eyeglasses category.
On novel real laptop trajectories from the BMVC dataset \cite{BMVC2015_181}, we achieve the best performance directly generalized from SAPIEN without further fine-tuning.
Finally, our extensive experiments further demonstrate the robustness of our tracking method to pose errors and achieve the fastest speed ($\sim 12$ FPS) among all methods.

\vspace{-3mm}
\section{Related Works}
\vspace{-2mm}
\noindent\textbf{Category-Level Object Pose Estimation}
To define category-level poses of novel object instances, Wang \emph{et al.} \cite{wang2019normalized} proposed Normalized Object Coordinate Space (NOCS) as a category-specific canonical reference frame for rigid objects. The objects from the same category in NOCS are consistently aligned to a category-level canonical orientation. These objects are further zero-centered and uniformly scaled so that their tight bounding boxes are centered at the origin of NOCS with a diagonal length of 1. 
Li \emph{et al.} \cite{li2020categorylevel} extended the definition of NOCS to rigid parts in articulated objects and proposed  Normalized Part Coordinate Space (NPCS), which is a part-level canonical reference frame(see appendix \ref{sec:nocs} for a detailed introduction).
Several works have been improving \cite{wang2019normalized} via leveraging analysis-by-synthesis and shape generative models as in \cite{chen2020category,chen2020learning} and learnable deformation as in \cite{tian2020shape}.
Most of these methods leverage RANSAC for pose fitting, which prohibits their pipelines from being end-to-end differentiable, potentially rendering those methods sub-optimal. Although several works have proposed differentiable RANSAC layers to bridge this gap, \textit{e.g.}, DSAC \cite{brachmann2017dsac}, DSAC++ \cite{brachmann2018learning}, we will show that our method performs better than these methods without using RANSAC.

\noindent\textbf{Category-Level Object Pose Tracking}
As the only existing work in this field, Wang \emph{et al.} \cite{wang20196} proposed a 6D Pose Anchor-based Category-level Keypoint tracker (6-PACK) by tracking keypoints in RGB-D videos. 6-PACK first employs an attention mechanism over anchors and then generates keypoints in an unsupervised manner, which are used to compute interframe pose changes. %
It is important to note several key differences between 6-PACK and our work: 
1) 6-PACK is designed only for rigid objects and cannot handle articulated objects;
2) 6-PACK only estimates the 6D pose containing rotation and translation and omits the important 3D size estimation required to obtain the 3D amodal object bounding boxes.

As special cases of category-level articulated object pose tracking, hand and human pose tracking problems are very popular due to their broad applications \cite{oikonomidis2011efficient, wang2009real, mueller2018ganerated, han2020megatrack, ganapathi2012real, xiao2018simple, andriluka2018posetrack}. However, the developed methods leverage domain-specific knowledge of hand and human body, thus prevent them from being applied to generic articulated objects.

\noindent\textbf{Instance-Level 6D Pose Tracking}
Instance-level pose tracking works track the poses of known object instances.
Classic methods, \textit{e.g.}, ICP-based tracking \cite{zhou2018open3d}, explicitly rely on the object CAD models. 
Some recent works \cite{choi2013rgb,deng2019poserbpf,wuthrich2013probabilistic,krull20146, deng2019poserbpf} use particle filtering to estimate the posterior of object poses. 
Other methods measure the discrepancy between the current observation and the previous state, and perform tracking via optimization \cite{schmidt2014dart,pauwels2015simtrack}.
The most relevant works to ours are delta pose based methods \cite{li2018deepim,wen2020se}, which perform tracking by regressing the pose change between consecutive frames using deep neural networks.
\vspace{-2mm}

\section{Problem Definition and Notations}
\begin{figure*}[t]
\begin{center}
   \includegraphics[width=\linewidth]{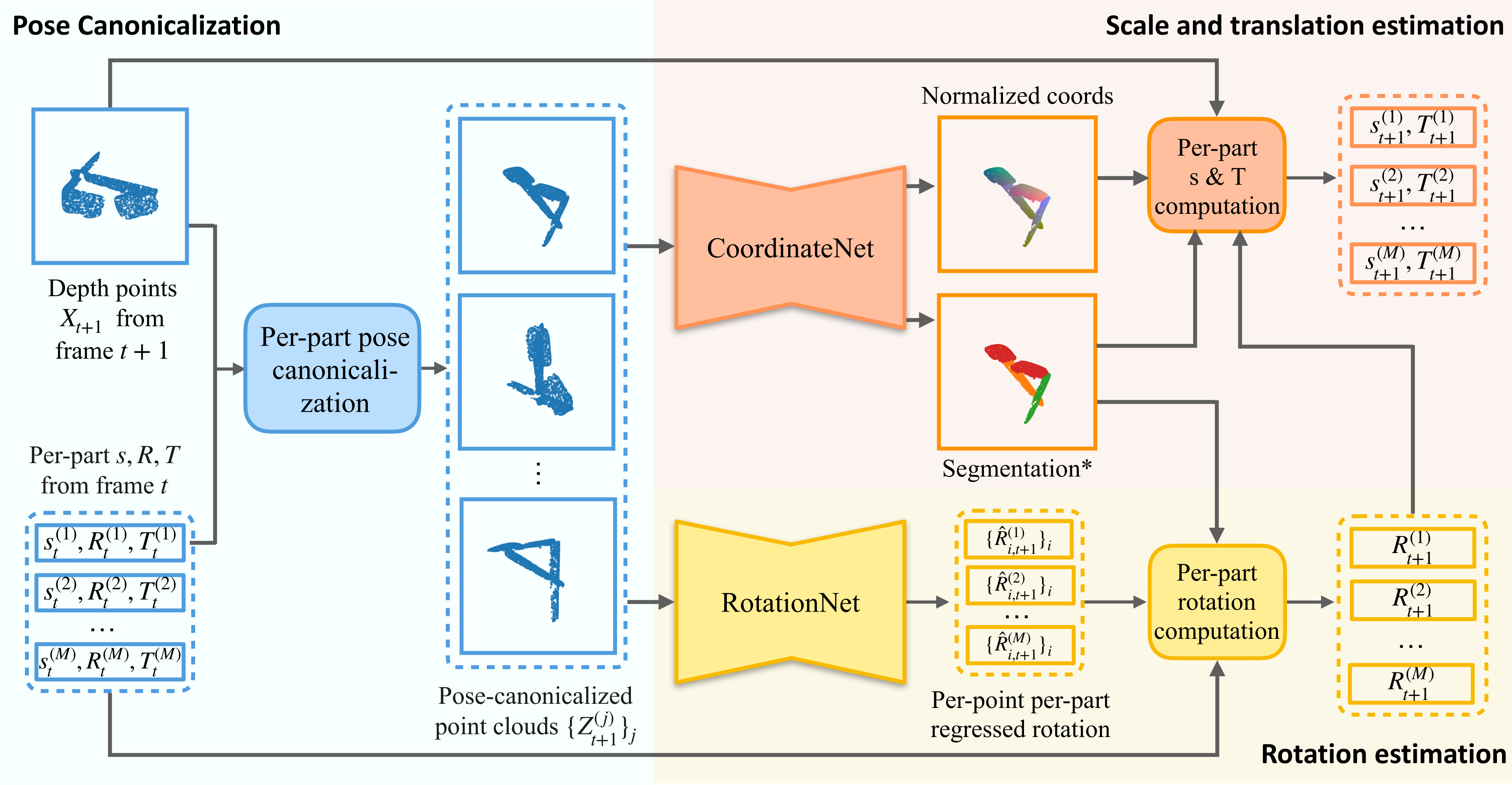}
\end{center}
   \vspace{-3mm}
   \caption{\textbf{Our end-to-end differentiable pose tracking pipeline} takes as inputs a depth point cloud of an $M$-part object along with its per-part scales, rotations, and translations estimated from the last frame. We first adopt per-part pose canonicalization to transform the depth points using the inverse estimated pose and generate $M$ pose canonicalized point clouds. The canonicalized point clouds will be fed into RotationNet for per-part rotation estimation, as well as CoordinateNet for part segmentation and normalized coordinate predictions, which are used to compute the updated scales and translations. When RGB images are available, segmentation can be replaced by the results from the off-the-shelf image detectors for better accuracy. Such a pipeline can be naturally adapted to rigid objects when $M=1$.
   } 
   \vspace{-5mm}
\end{figure*}

\vspace{-2mm}
In this paper, we target at the problem of tracking the 9DoF poses of rigid or articulated objects from known categories. We follow the category-level rigid object and part pose definition in \cite{wang2019normalized,li2020categorylevel} and adopt the assumption in \cite{li2020categorylevel} that the number of rigid parts $M$ is known and constant for all the objects in a known category, where $M>1$ indicates an articulated object category, and $M=1$ indicates a rigid object category. Without loss of generality, we only describe the notations for articulated object pose tracking. For a point cloud $X = \{x_i \in \mathbb{R}^{3}\}_{i=1}^N$ containing object instance $O = \{C^{(j)}\}_{j=1}^M$, where $N$ is the number of points and $C^{(j)}\subset X$ represents points of the $j$-th rigid part, we denote category-level part pose as $\mathtt{P}^{(j)} = \{d^{(j)}, R^{(j)}, T^{(j)}\}$, where $d^{(j)}\in \mathbb{R}^3$ is 3D size, $R^{(j)} \in SO(3)$ is rotation, and $T^{(j)} \in \mathbb{R}^3$ is translation.

Our problem is then defined as follows:
Given a live stream of depth point clouds $\{X_t\}_{t\geq 0}$ containing object instance $O$ along with its per-part pose initialization $\{\mathtt{P}^{(j)}_0\}_{j}$, our objective is to track its part poses $\{\mathtt{P}^{(j)}_t\}_{j}$ in an online manner, where we process the point clouds and estimate the poses for all the frames $t>0$. In other words, at frame $t+1$, given the estimated $\{\mathtt{P}^{(j)}_t\}_{j}$ from frame $t$ and the depth point cloud $X_{t+1}$, our system needs to estimate $\{\mathtt{P}^{(j)}_{t+1}\}_{j}$.

\section{End-to-end Differentiable Pose Tracking}

\vspace{-1mm}
In this section, we introduce our approach in detail. We present the pose canonicalization module in Section \ref{sec:canon}, and pose tracking in Section \ref{sec:diff}, which includes the proposed RotationNet module and CoordinateNet module, finally, we describe our training protocol in Section \ref{sec:training}. The entire framework is differentiable and end-to-end trained, without any pre- or post- processing.
\vspace{-1mm}
\subsection{Pose Canonicalization} \label{sec:canon}
\vspace{-1mm}
Inspired by \cite{wang2019normalized}, we factorize the 9DoF pose $\mathtt{P}^{(j)}$ prediction  into a 7DoF similarity transformation ${\mathcal T}_t^{(j)}\in \text{Sim}(3)$ estimation problem and a 3D aspect ratio $\widehat{d}^{(j)}$ estimation problem.  
Formally, we define the per-part 1D uniform scale as $s^{(j)} = ||d^{(j)}||$ and 3D aspect ratio as $\hat{d^{(j)}} = d^{(j)} / s^{(j)}$ so that $d^{(j)} = s^{(j)}\hat{d^{(j)}}$. 
We can then represent $\mathcal{T}^{(j)} = \{s^{(j)}, R^{(j)}, T^{(j)}\} $. 

To estimate $\mathcal{T}^{(j)}_{t+1}$ from the observed point cloud $X_{t+1}$, there are two types of approaches. One way is to train a neural network for direct pose regression. 
Another way is to estimate the normalized coordinates $Y^{(j)}_{t+1}$ of $C^{(j)}_{t+1}$, which satisfy $C^{(j)}_{t+1} = s^{(j)}_{t+1}R^{(j)}_{t+1}Y^{(j)}_{t+1} + T^{(j)}_{t+1}$, and then compute the $\mathcal{T}^{(j)}_{t+1}$ 
using the Umeyama algorithm \cite{umeyama1991least} along with RANSAC, 
thus the 3D aspect ratios $\hat{d^{(j)}}$ can be estimated using the axis range $(|x|_{max}, |y|_{max}, |z|_{max})$ of $Y^{(j)}_{t+1}$.

In our framework, to simplify the learning problem of mapping the input $X_{t+1}$ to the output $\mathcal{T}^{(j)}_{t+1}$, we propose to canonicalize both its input and output spaces using $\mathcal{T}^{(j)}_{t}$, which allows to further combine the two aforementioned methods.
\vspace{-2mm}

\begin{definition} [Pose-canonicalized point cloud]
Pose-canonicalized point cloud $Z_{t+1}^{(j)}$ with respect to part $j$ in the input point cloud $X_{t + 1}$ is defined as the product of the inverse transformation of $\mathcal{T}_{t}^{(j)}$ and $X_{t+1}$, namely $Z_{t+1}^{(j)} =  \left(R_t^{(j)}\right)^{-1}\left(X_{t+1} - T_t^{(j)}\right)/s_t^{(j)}$.
\end{definition}
\vspace{-2mm}

We observe that for the learning problem that maps $X_{t+1}$ to $\mathcal{T}^{(j)}_{t+1}$, by canonicalizing its input $X_{t+1}$ to pose-canonicalized point cloud $Z_{t+1}^{(j)}$, its output $\mathcal{T}^{(j)}_{t+1}$ will be canonicalized to $\hat{\mathcal{T}}^{(j)}_{t+1} = \{\hat{s}^{(j)}_{t+1}, \hat{R}^{(j)}_{t+1}, \hat{T}^{(j)}_{t+1}\}$ correspondingly, where $\hat{s}^{(j)}_{t+1} \approx 1, \hat{R}^{(j)}_{t+1} \approx I, \hat{T}^{(j)}_{t+1} \approx 0$. (See appendix \ref{sec:proof} for the proof.)

Note that $\mathcal{T}^{(j)}_{t+1}$ can be expressed using $\hat{\mathcal{T}}_{t + 1}^{(j)}$ and $\mathcal{T}^{(j)}_{t}$, namely, $s_{t+1}=s_t^{(j)}\hat{s}_t^{(j)}$, $R_{t + 1}^{(j)}=R_t^{(j)}\hat{R}_t^{(j)}$, $T_{t + 1}^{(j)}=s_t^{(j)}R_t^{(j)}\hat{T}_{t}^{(j)} + T_t^{(j)}$.
Now the pose estimation problem has been transformed and canonicalized to learning a mapping from $Z_{t + 1}^{(j)}$ to $\hat{\mathcal{T}}_{t + 1}^{(j)}$. The input $Z_{t + 1}^{(j)}$ contains $\hat{C}_{t+1}^{(j)} = \left(R_t^{(j)}\right)^{-1}\left(C_{t+1}^{(j)} - T_t^{(j)}\right)/s_t^{(j)}$ that aligns well across different frames and the output space is quite constrained around an identity transformation. In this way, we simultaneously canonicalize the input point cloud space and the output pose space. By doing so, we significantly simplify the regression task, yielding much improved pose estimation accuracy and better generalizability across different instances. 
Note that turning the estimation of ${\mathcal{T}}_{t + 1}^{(j)}$ into $\hat{\mathcal{T}}_{t + 1}^{(j)}$ is closely related to estimating the interframe delta 6D pose widely used in instance-level 6D pose estimation and tracking works \cite{li2018deepim,wen2020se}. 
To be more specific, we estimate interframe 7D delta transformations in a category-level canonical frame, \textit{i.e.}, NOCS for rigid objects and NPCS for parts, whereas none of the input and output spaces of delta pose estimation in \cite{li2018deepim,wen2020se} is canonical.

\vspace{-2mm}
\subsection{Pose Tracking} \label{sec:diff}
\vspace{-2mm}
Taking the pose-canonicalized point cloud $Z_{t + 1}^{(j)}$ as input, we learn a RotationNet (see Section \ref{sec:rotation}) that directly regresses $\hat{R}_{t+1}^{(j)}$ and then recovers ${R}_{t+1}^{(j)} = {R}_{t}^{(j)}\hat{R}_{t+1}^{(j)}$. In contrast to regressing arbitrary rotations in nonlinear $SO(3)$ space, since $\hat{R}_{t+1}^{(j)}$ lies in a small, near-linear neighborhood around $I_{3 \times 3}$, the regression can be done with high accuracy (see appendix \ref{sec:pca} for more details).
However, we experimentally find directly regressing $\hat{s}_t^{(j)}$ and $\hat{T}_t^{(j)}$ still difficult, due to the translation ambiguity caused by incompleteness of the partial observations $Z_{t+1}^{(j)}$. Imagine a pencil with one end occluded, the length of the pencil cannot be determined, thus making prediction of its center translation highly ambiguous. 
Although certain cues, \textit{e.g.}, object symmetry, may help relieve the ambiguity, our experiments show that given partial observations of asymmetric objects, regressing $\hat{T}_t^{(j)}$ still remains challenging. In contrast, our experiments reveal that predicting normalized coordinates $Y_{t+1}^{(j)}$ from $Z_{t+1}^{(j)}$ is quite successful on all rigid and articulated objects, which implicitly estimates $\hat{s}_t^{(j)}$ and $\hat{T}_t^{(j)}$. 
The reason for this success is that the normalized coordinates $Y_{t+1}^{(j)}$ capture the category-wise prior and enforces a zero-centered frame, thus making the translation estimation more well-considered and accurate than direct regression. We therefore devise a CoordinateNet to segment $C^{(j)}_{t+1}$ from $X^{(j)}_{t+1}$ and predict $Y_{t+1}^{(j)}$ (see Section \ref{sec:scale}).

By combining the RotationNet and CoordinateNet's outputs and knowing $C^{(j)}_{t+1} = s^{(j)}_{t+1}R^{(j)}_{t+1}Y^{(j)}_{t+1} + T^{(j)}_{t+1}$, we can analytically compute ${s}_{t+1}^{(j)}$ and ${T}_{t+1}^{(j)}$ via the Umeyama algorithm \cite{umeyama1991least}
(assume $R$ is given). Usually a non-differentiable RANSAC is needed when using Umeyama algorithm as in \cite{wang2019normalized, li2020categorylevel} due to the multi-modal noises in the predicted $Y_{t+1}^{(j)}$. Thanks to the pose canonicalization, we find that our  $Y_{t+1}^{(j)}$ predictions are very successful and RANSAC only brings limited improvement to our predictions (see Sec. \ref{sec:robustness}).

Being free from the non-differentiable RANSAC step, our end-to-end differentiable pose tracking pipeline can be straightly optimized for pose accuracy, enforce pose losses (\textit{e.g.}, IoU loss) directly at its outputs (see Section \ref{sec:training}), and improve its running speed.
\vspace{-3mm}
\subsubsection{Rotation Estimation} \label{sec:rotation}
\vspace{-2mm}
\paragraph{RotationNet}
To predict $\{\hat{R}_{t+1}^{(j)}\}_j$ for each individual part, we devise a point cloud based deep neural network, RotationNet, that takes as inputs the pose-canonicalized point clouds $\{Z_{t+1}^{(j)}\}_j$ with respect to each individual part $j$. Built upon PointNet++ \cite{qi2017pointnet++}, RotationNet regresses per-point per-part rotations $\{\hat{R}_{i,t}^{(j)}\}_{i,j}$ in the form of the 6D continuous rotation representation  \cite{zhou2019continuity}. Note that the PointNet++ is not deterministic since it uses random further point sampling in both set abstraction and ball query operations, thus resulting in difficulties achieving convergence on accurate regression tasks. To suppress noise, we average across the rotation predictions using the Euclidean mean \cite{moakher2002means} from points on part $j$ to obtain the final prediction $\hat{R}_{t+1}^{(j)}$. 

For symmetric objects such as bowls, rotation ambiguity exists around their symmetric axis. See appendix \ref{sec:sym_rotation} for how we supervise rotation for them.

\noindent\textbf{Training and Inference}
At training time, we enforce a per-point mean square loss for points inside the ground truth mask $m_{t+1}^{(j)}$. At test time, the mask comes from the predicted part segmentation from CoordinateNet.

\vspace{-5mm}
\subsubsection{Scale and Translation Estimation} \label{sec:scale}
\vspace{-2mm}
\noindent\textbf{CoordinateNet}
To estimate $\{Y_{t+1}^{(j)}\}_j$, we devise CoordinateNet that takes as input the pose-canonicalized point cloud $Z_{t+1}^{(1)}$ with respect to the first part ($j=1$) and predicts its per-point part segmentation and per-point per-part normalized coordinates $\{Y_{i,t+1}^{(j)}\}_{i,j}$. 
Note that pose-canonicalized point clouds with respect to different parts share the same segmentation and normalized coordinates; thus, we only need to take $Z_{t+1}^{(1)}$ as the input.

Built upon a PointNet++ segmentation network, CoordinateNet branches into two heads after the final feature propagation layers: one head for segmentation and the other for normalized coordinate prediction. For the segmentation head, we use relaxed IoU loss \cite{yi2018deep}. For the normalized coordinate head, we predict class-aware normalized coordinates, similar to \cite{wang2019normalized, li2020categorylevel}.
During training, we enforce an RMSE loss on the points inside the ground truth part masks. At test time, we use predicted masks to select coordinate predictions from $M$ parts.

For symmetric objects, see appendix \ref{sec:sym_coord} for how we handle ambiguity in their normalized coordinates.

\noindent\textbf{Per-part Scale and Translation Computing}
Knowing the dense correspondence between $Y_{t+1}^{(j)}$ and $C_{t+1}^{(j)}$ and assuming $R_{t+1}^{(j)}$ is given by RotationNet, we can analytically compute ${s}_{t+1}^{(j)}$ and ${T}_{t+1}^{(j)}$ via the Umeyama algorithm \cite{umeyama1991least}.
See appendix \ref{sec:compute_st} for further details.

\vspace{-1mm}
\subsection{Training Protocol} \label{sec:training}
\vspace{-1mm}
\noindent\textbf{Training Data Generation}
To train CoordinateNet and RotationNet, we need paired data of pose-canonicalized point clouds and their corresponding ground truth poses. We propose to generate the training data on-the-fly without using any real temporal data.
For a depth point cloud $X$ and part $j$ in it, we perturb its per-part ground truth scale, rotation, and translation by adding random Gaussian noise to them, namely $s'^{(j)} = s^{(j)} (1+n_s)$, $R'^{(j)} = R^{(j)}R_\text{rand}$, $T'^{(j)} = T^{(j)} + n_T$, where $n_s \sim \mathcal{N}(0, \sigma_s)$, $R_\text{rand}$ is a rotation matrix with a random axis and a random angle $n_{\theta} \sim \mathcal{N}(0, \sigma_{\theta})$, and $n_T$ is a 3D vector with a random direction and a random length $t \sim \mathcal{N}(0, \sigma_{T})$. We then generate the pose-canonicalized point clouds and compute their corresponding ground truth. 

\noindent\textbf{Pose Losses for RotationNet and CoordinateNet}
For $s_{t+1}^{(j)}$, $R_{t+1}^{(j)}$ , $T_{t+1}^{(j)}$, their predictions are end-to-end differentiable; we thus propose to additionally enforce pose losses directly on these predictions. We use RMSE loss for supervising scale error $L_{scale}$ and translation error $L_{trans}$. To directly improve the final 3D IoU predictions, we incorporate a corner loss $L_{corner}$ \cite{manhardt2019roi}, defined as the corresponding per-vertex distances between the ground truth bounding box in the camera frame and the ground truth bounding box in the normalized coordinate space transformed by our predicted $s_{t+1}^{(j)}$, $R_{t+1}^{(j)}$ , $T_{t+1}^{(j)}$. For symmetric objects, we enforce the corner loss on the two intersection points of the symmetric axis and the bounding box surface. The total loss $L_\text{total} = \lambda_\text{seg}L_\text{seg} + \lambda_\text{coord}L_\text{coord} + \lambda_\text{rot}L_\text{rot} + \lambda_\text{scale}L_\text{scale} + \lambda_\text{translation}L_\text{translation} + \lambda_\text{corner}L_\text{corner}$.

\vspace{-2mm}
\section{Experiment}
\begin{center}
\begin{table*}[t]
\centering
\setlength{\tabcolsep}{3pt}
\resizebox{0.85\textwidth}{!}{
\begin{tabular}{|c|c|c|c|c|c|c|c|c|}
\hline
{Method}          & NOCS\cite{wang2019normalized}  & CASS\cite{chen2020learning}  & CPS++\cite{manhardt2020cps} & Oracle ICP & 6-PACK\cite{wang20196}    & 6-PACK \cite{wang20196}    & Ours & Ours+RGB seg. \\ \hline
{Input}           & RGBD  & RGBD  & RGB & Depth & RGBD & RGBD & Depth & RGBD \\\hline 
{Setting}         & \multicolumn{3}{|c|}{Single frame} & \multicolumn{5}{|c|}{Tracking} \\ \hline
{Initialization}           & N/A   & N/A   &  N/A  & GT. & GT.  & Pert.  & Pert. & Pert. \\ \hline
{5$^{\circ}$5cm $\uparrow$} & 16.97 & 29.44 &  2.24 & 0.65 & 28.92 & 22.13 & {62.16} & \textbf{63.60} \\ 
{mIoU$\uparrow$}           & 55.15 & 55.98 & 30.02 & 14.69 & 55.42 &53.58 & {64.10} & \textbf{69.19} \\ 
{$R_{err}\downarrow$}      & 20.18 & 14.17 & 25.32 & 40.28 & 19.33 & 19.66 & \textbf{5.94} & 6.43 \\ 
{$T_{err}\downarrow$}      &  4.85 & 12.07 & 21.62 & 7.71 & \textbf{3.31} & 3.62 & 7.92 & 4.18 \\ \hline
\end{tabular}
}
\vspace{1mm}
\caption{
\textbf{Results of category-level rigid object pose tracking on NOCS-REAL275}. The results are averaged over all 6 categories.}
\label{table:rigid}
\vspace{-4mm}
\end{table*} 
\end{center}

\vspace{-10mm}
\subsection{Datasets and Evaluation Metrics}
\vspace{-1mm}
\paragraph{NOCS-REAL275}
For rigid object pose tracking, we evaluate our methods on the NOCS dataset \cite{wang2019normalized} that contains six categories: bottle, bowl, camera, can, laptop, and mug, where bottle, bowl, and can are symmetric.
The training set contains: 1) the train split of the CAMERA dataset \cite{wang2019normalized}, composed of 300K mixed reality data with synthetic object models from ShapeNetCore \cite{chang2015shapenet} as foregrounds and real backgrounds captured in IKEA; and 2) seven real videos capturing challenging cluttered scenes with three object instances in total for each object category. The testing set, NOCS-REAL275, has six real videos depicting in total three different (unseen) instances for each object category totaling 3200 frames.

\noindent\textbf{Articulated Objects from SAPIEN}
For articulated object pose tracking, we create a synthetic dataset using SAPIEN \cite{xiang2020sapien}. Our dataset contains four categories: glasses, scissors, laptop, and drawers, where drawers have prismatic joints and the others have revolute joints. The training set contains 98K depth images of 164 standalone object instances with random joint states and viewpoints. The testing set contains 190 depth videos of 19 unseen instances with a total length of 19K frames, where instances keep moving and changing their joint states. See appendix \ref{sec:sapien_dataset} for more information.

\noindent\textbf{Real-World Laptop Test Trajectories from the BMVC dataset \cite{BMVC2015_181}} 
We also test our model on real articulated object trajectories from \cite{BMVC2015_181}, where the objects maintain the same joint state and only viewpoint changes.
Among the 4 instances in the dataset, we can only evaluate our method on the laptop for which we have category-level training data from SAPIEN. 
The two laptop depth sequences contain a total of 1765 frames. 

\noindent\textbf{Evaluation Metrics}
We report the following metrics for both rigid and articulated object pose tracking: 1) \textbf{5$^{\circ}$5cm accuracy}, the percentage of pose predictions with rotation error $<$ 5$^{\circ}$ and translation error $<$ 5cm; 2) \textbf{mIoU}, the average 3D intersection over union of ground-truth and predicted bounding boxes; 3) {\boldmath{$R_{err}$}}($^{\circ}$), average rotation error; and 4) {\boldmath{$T_{err}$}}(cm), average translation error. For articulated objects, we additionally report the average joint state accuracy: 5) {\boldmath{$\theta_{err}$}}($^{\circ}$) rotation error for revolute joints; and 6) {\boldmath{$d_{err}$}}(cm) translation error for prismatic joints. For real-world laptop trajectories, we follow \cite{BMVC2015_181} and use pose tolerance, namely the Averaged Distance (AD) accuracy with ${10\%}$ of the object part diameter as the threshold.

\vspace{-1mm}
\subsection{Category-Level Rigid Object Pose Tracking}\label{sec:exp_rigid}
\vspace{-1mm}
\noindent\textbf{Experiment Setting} To track an object in the cluttered scenes from the NOCS-REAL275 dataset, we propose to first crop from the scene a ball of depth points enclosing the object of interest. We set the center and the radius of the ball according to the previous frame's 9DoF pose estimation. To generate training data, we jitter the ground-truth pose with Gaussian noises ($\sigma_{scale} = 0.02$, $\sigma_{rot} = 5^{\circ}$, and $\sigma_{trans} = 3 \text{cm}$) to mimic interframe pose changes and crop balls accordingly. 
At test time, we randomly sample an initial pose around the ground-truth pose for the first frame from the same distribution. 

\noindent\textbf{Results} Table \ref{table:rigid} summarizes the quantitative results for rigid object pose tracking. We report the performance of our method using only depth and using RGBD where
object segmentation masks can be obtained by running off-the-shelf detectors on RGB images as in CASS\cite{chen2020learning}.
We compare our method with: {6-PACK} \cite{wang20196}, a tracking based method that is initialized with the same pose error distributions or ground-truth poses (6-PACK originally only initializes with translation errors); Oracle ICP, which leverages the ground truth object models;
and several single-frame based method, including {NOCS} \cite{wang2019normalized}, {CASS} \cite{chen2020learning} and {CPS++} \cite{manhardt2020cps}.

Our method achieves the best performance and significantly outperforms the previous state-of-the-art method, 6-PACK, under both initialization settings. We are especially competitive under the rotation error and 5$^{\circ}$5cm metrics, showing less than a third of the rotation error and a 105\% higher 5$^{\circ}$5cm percentage compared to 6-PACK. Using only depth, our method generates relatively weaker performance regarding translation error, however, this is only due to the failure to segment out cameras on the real test depth images, given the huge sim2real domain gap between our mostly synthetic training data and noisy real data. See section \ref{sec:failure_case} and appendix \ref{sec:per_category_rigid} for detailed analysis.
Excluding this camera category, {our} method will be the best under all metrics (see appendix \ref{sec:per_category_rigid}).
It is worth noting that while our method tracks the full 9DoF pose and predicts the bounding boxes, 6-PACK only tracks the 6DoF rigid transformation and has to use a ground-truth box scale to compute 3D IoU, which unfairly advantages 6-PACK during the comparison.
Fig.\ref{fig:teaser} further shows the qualitative comparison between our method and 6-PACK. Our method loses track less often and gives better pose estimations. 

\begin{center}
\begin{table}[t]
\setlength{\tabcolsep}{4pt}
 \resizebox{0.48\textwidth}{!}{
      \centering
\begin{tabular}{|c|c|c|c|c|c|c|}
\hline
Method  & 5$^{\circ}$ 5cm$\uparrow$ & mIoU$\uparrow$  & $R_{err}\downarrow$ & $T_{err}\downarrow$ & $\theta_{err}\downarrow$ & $d_{err}\downarrow$\\ \hline
 ANCSH* \cite{li2020categorylevel}             & 92.55         & 	68.69       &	 2.18 &  0.48 & 1.62 & 0.64 \\
 Oracle ICP          & 62.87         &   56.61       &	8.95 &	 3.04 & 7.21 & 1.05 \\ 
 Ours               & \textbf{98.35}& \textbf{74.00}&	\textbf{1.03} &	\textbf{0.29} & \textbf{1.38}& \textbf{0.34} \\
 \hline\hline
 \textit{C}-sRT regression              & 21.69         &	34.21       &	20.48 &	11.46 & 6.08 & 7.57 \\
 \textit{C}-CoordinateNet             & 95.06         &	71.99       &	 2.09 &	 0.40 & 1.52 & 0.75\\ 
\textit{C}-Crd. + DSAC++ \cite{brachmann2018learning}   & 95.68         &   68.21       &	 1.80 &	 0.47 & 1.61 & 0.56 \\ 
 Ours w/o $L_{c}, L_{s}, L_{t}$ 
                    & 97.63         &	72.09       &	 1.24 &	 0.35 & 1.43 & 0.36\\ \hline\hline
  Ours + Rot. Proj.   & 98.74         &   74.17       &    0.97 & 0.29 & 1.37 & 0.34\\

\hline
\end{tabular}
}
\vspace{0.2mm}
\caption{\textbf{Experiment results and ablation studies of articulated object pose tracking on the held-out instances from SAPIEN.} $\theta_{err}$ is averaged over all revolute joints, while $d_{err}$ is averaged over all prismatic joints. Other results are averaged over parts and categories. See appendix \ref{sec:per_category_articulated} for per-part, per-category results. {Ours + Rot. Proj} leverages kinematic constraints, see Section \ref{sec:discussion}.}
\label{table:articulated}
\vspace{-2mm}
\end{table}
\end{center}

\vspace{-10mm}
\subsection{Category-Level Articulated Pose Tracking}\label{sec:exp_art}
\vspace{-1mm}
In Table \ref{table:articulated} and Fig.\ref{fig:teaser}, we show our articulated pose tracking results on held-out test instances from the SAPIEN dataset. 
We compare our method to 1) {ANCSH*} (oracle ANCSH), where we provide ground-truth object segmentation masks to the state-of-the-art single frame articulated object pose estimation work, ANCSH \cite{li2020categorylevel}.
The original ANCSH fails drastically on part segmentation on our dataset due to the part ambiguity of textureless object point clouds rendered from arbitrary viewpoints; 
and 2) {oracle ICP}, where we assume available ground-truth part labels and object part models, then track each part using ICP.

Note that our articulated SAPIEN dataset is depth-only while RGB-D input is essential for 6-PACK, we thus did not run per-part 6-PACK tracking as a baseline. 

We outperform the baselines under all metrics. Although {ANCSH*} uses ground-truth labels and regulates its predictions with joint constraints, our per-part scheme still beats it with exceptionally precise rotation estimations.

\begin{center}
\begin{table}[t]
\setlength{\tabcolsep}{3pt}
 \resizebox{0.48\textwidth}{!}{
      \centering
\begin{tabular}{|c|c|c|c|c|c|}
\hline

\multicolumn{2}{|c|}{Method} & Michel et al. & ANCSH         & ANCSH* & Ours           \\ \hline
\multicolumn{2}{|c|}{Setting} & \multicolumn{2}{|c|} {Known instance} & \multicolumn{2}{|c|}{Category-level}          \\\hline
1   & all / parts     & 64.8 / 65.5 66.9 & 94.1 / 97.5 94.7 &  74.7 / 89.1 78.5 & \textbf{95.5} / \textbf{99.8 95.7}         \\\hline
2   & all / parts    & 65.7 / 66.3 66.6 & 98.4 / 98.9 \textbf{99.0} & 97.0 / 98.0 97.6 & \textbf{98.9} / \textbf{100.0} 98.9 \\ \hline
\end{tabular}
}
\vspace{0.2mm}
\caption{\textbf{Results on two real sequences of an unseen laptop} are measured in pose tolerance (the higher, the better, see \cite{BMVC2015_181}). The left two columns reported by \cite{BMVC2015_181, li2020categorylevel} are directly trained on the instance, whereas {ANCSH*}(with GT part mask) and ours are only trained on SAPIEN and have never seen the instance.}
\label{table:bmvc}
\vspace{-2mm}
\end{table}
\end{center}

\vspace{-9mm}
\subsection{Category-Level Articulated Pose Tracking on Real-World Data}\label{sec:exp_art_real}
\vspace{-2mm}
We further test our model, trained on the synthetic SAPIEN dataset only, on real-world data. Since the training data does not contain backgrounds, we use pre-segmented object point clouds in the following experiments. 

\noindent\textbf{Real Laptop Trajectories} In Table \ref{table:bmvc} and Fig.\ref{fig:teaser}, we compare our method to Michel et al. \cite{BMVC2015_181}, {ANCSH} \cite{li2020categorylevel}, and {ANCSH*} on two real laptop trajectories from \cite{BMVC2015_181}. We follow \cite{li2020categorylevel} and use their rendered object masks for segmentation. In spite of a Sim2Real gap and a category-level generalization gap, our model outperforms all other methods.

\begin{figure}[t]
\begin{center}
   \includegraphics[width=1.0\linewidth]{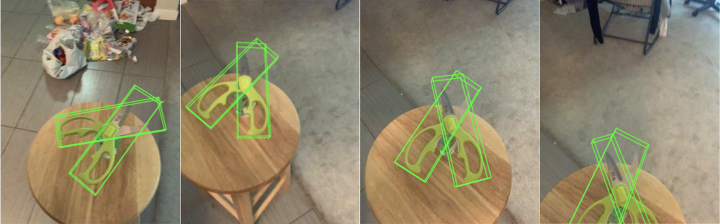}
\end{center}
    \vspace{-4mm}
   \caption{\textbf{Qualitative evaluation on the real scissors trajectory.}}
   \vspace{-1mm}
   \label{fig:real_scissors}
\end{figure}

\noindent\textbf{Real Drawers Under Robot-Object Interaction} To test our method in robotic manipulation scenarios, we capture a real drawers trajectory using Kinect2, where a Kinova Jaco2 arm pulls out the middle drawer. As shown in Fig.\ref{fig:teaser}(d). Our model successfully tracks both the moving drawer and the other parts. See appendix \ref{sec:real_art_detail} for more details.

\noindent\textbf{Real Scissors Trajectories} Fig.\ref{fig:real_scissors} shows quantitative results on a real scissors trajectory we captured using Kinect2.

\begin{center}
\begin{table}[t]
\setlength{\tabcolsep}{3pt}
\resizebox{0.5\textwidth}{!}{
\begin{minipage}{0.55\textwidth}
\begin{tabular}{|c|c|c|c|c|c|c|}
\hline

\multirow{2}{*}{Method}  & \multirow{2}{*}{CrdNet}  & \multirow{2}{*}{$C$-Crd.}  & \multirow{2}{*}{\shortstack{$C$-Crd.+ \\ DSAC++}} & \multirow{2}{*}{$C$-sRT} & \multirow{2}{*}{\shortstack{Ours w/o \\$L_c, L_s, L_t$}}  & \multirow{2}{*}{Ours} \\ 
       &               &         &       &       &               & \\ \hline
 
 5$^{\circ}$5cm $\uparrow$ & {14.93} & {46.74} & 54.77         & 25.99 & {60.48} & \textbf{{62.16}} \\
 mIoU$\uparrow$            & {49.48} & {59.99} & 53.89         & 32.86 & 58.80          & \textbf{{64.10}} \\
 $R_{err}\downarrow$       & {53.63} & {35.08} & {8.88} & 34.74 & {6.41}  & \textbf{{5.94}}  \\
 $T_{err}\downarrow$       & {9.48}  & {12.97} & {9.95} & 21.84 & {12.64} & \textbf{{7.92}}  \\ \hline
\end{tabular}
\end{minipage}
}
\vspace{1mm}
\caption{\textbf{Ablation study of rigid object pose tracking on NOCS-REAL275.} The results are averaged over all 6 categories. Here \textit{C} represents canonicalized.}
\vspace{-5mm}
\label{table:ablation}
\end{table}
\end{center}

\vspace{-12mm}
\subsection{Ablation Study}
\vspace{-1mm}
To demonstrate the effectiveness of our multi-component design, we construct several variants of our network: 
1) {CoordinateNet}, where we directly regress the NOCS/NPCS coordinates from $X$ without pose canonicalization;
2) {canonicalized CoordinateNet}, where we perform pose canonicalization but don't have RotationNet;
3) {canonicalized CoordinateNet with DSAC++}, where we follow \cite{brachmann2018learning} and train our CoordinateNet with a differentiable pose estimation module; 
4) {canonicalized sRT regression}, where we extend our RotationNet to further regress scale and translation based on canonicalized point clouds without using CoordinateNet;
and 5) {Ours w/o} {$L_c, L_s, L_t$ losses}, where we discard the pose losses $L_{scale}, L_{trans}, L_{corner}$ during training. 
For 1), 2), and 3) we take the coordinate predictions from CoordinateNet and use RANSAC-based pose fitting.

We test the variants on NOCS-REAL275 for rigid object tracking and SAPIEN synthetic dataset for articulated object tracking. The results are summarized in Table \ref{table:ablation} and Table \ref{table:articulated}, where our method outperforms all ablated versions by successfully combining the benefits of pose canonicalization, coordinate prediction, and pose regression. Note that we did not test CoordinateNet without canonicalization on articulated objects due to the part ambiguity of uncolored, arbitrarily posed synthetic objects.

{Canonicalized CoordinateNet} significantly outperforms {CoordinateNet}, demonstrating the benefit brought by pose canonicalization. 
With additional RotationNet, our method further improves {canonicalized CoordinateNet} and beats the differentiable pipeline {CoordinateNet + DSAC++} which also includes explicit pose losses, proving direct regression of small $\hat{R}_{t+1}^{(j)}$ to be a better choice in the tracking scenario. In contrast, due to ambiguities and insufficient visual cues about scale and translation in the input, the pure regression pipeline, {canonicalzied sRT regression}, produces the worst results. Finally, explicit scale, translation, and corner losses effectively improve our performance compared to ours w/o $L_c, L_s, L_t$ losses.

\begin{figure}
\centering
\begin{tikzpicture}
\begin{axis}[
    width=0.5\textwidth,
    height=0.4\textwidth,
    scale=1.0, 
    xmin=-0.5, xmax=2.5,
    ymin=0, ymax=70,
    xtick={0,1,2},
    xticklabels = {{Orig.}, {$+1\times$Noise}, {$+2\times$Noise}},
    ytick={0,15,30,45,60},
    legend style={at={(0.5, 0.45)}, 
    anchor=south, legend columns=2},
    ymajorgrids=true,
    point meta = y,
    grid style=dashed,
]

\addplot[
    color=blue,
    mark=*,
    ]
    coordinates {
    (0,22.13)(1,14.82)(2,7.44)
    };
    \node [above, color=blue] at (axis cs:  0,22.13) {\small $22.13$};
    \node [below, color=blue] at (axis cs:  1,14.82) {\small $14.82$};
    \node [above, color=blue] at (axis cs:  2, 7.44) {\small  $7.44$};
    \addlegendentry{\small 6-PACK, Init.}
    
\addplot[
    color=red,
    mark=*,
    ]
    coordinates {
    (0,62.16)(1,59.64)(2,55.94)
    };
    \node [below, color=red] at (axis cs:  1,59.64) {\small $59.64$};
    \node [below, color=red] at (axis cs:  2,55.94) {\small $55.94$};
    \addlegendentry{\small Ours, Init.}
    
\addplot[
    dashed,
    color=blue,
    mark=*,
    ]
    coordinates {
    (0,22.13)(1,17.95)(2,4.98)
    };
    \node [above, color=blue] at (axis cs:  1,17.95) {\small $17.95$};
    \node [below, color=blue] at (axis cs:  2,5.28) {\small $4.98$};
    \addlegendentry{\small 6-PACK, All}
    
\addplot[
    dashed,
    color=red,
    mark=*,
    ]
    coordinates {
    (0,62.16)(1,59.83)(2,58.69)
    };
    \node [above, color=red] at (axis cs:  0,62.16) {\small $62.16$};
    \node [above, color=red] at (axis cs:  1,59.83) {\small $59.83$};
    \node [above, color=red] at (axis cs:  2,58.69) {\small $58.69$};
    \addlegendentry{\small Ours, All}
    
\end{axis}
\end{tikzpicture}
\caption{\textbf{5$^{\circ}$5cm (\%) w.r.t. Additional Noise.} $+m\times$Noise means adding $m$ times train-time errors to 1) the initial pose (denoted \textbf{Init.}), which already contains $1\times$ train-time error; or 2) every frame during training (denoted \textbf{All}).}
\label{fig:robustness}
\vspace{-2mm}
\end{figure}
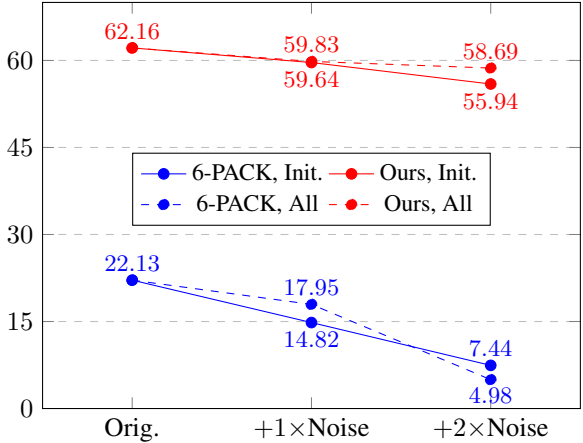

\vspace{-1mm}
\subsection{Tracking Robustness}\label{sec:robustness}
\vspace{-1mm}

Our pose prediction is conditioned on the pose from the previous frame, either the initial pose or an estimated pose. It is therefore worth testing the tracking robustness of our method against noisy pose inputs. As described in Sec. \ref{sec:exp_rigid} and Sec. \ref{sec:exp_art}, the initial pose errors are randomly sampled from Gaussian distributions. We directly test our model with 1 or 2 times of original pose errors added to (1) the initial pose, and (2) every previous frame's prediction, to examine the robustness to pose initialization and estimation errors, respectively. We plot the degradation of 5$^{\circ}$5cm accuracy under each setting and compare to 6-PACK in Fig.\ref{fig:robustness}. Our method is significantly more robust to pose noises than 6-PACK. We are also very robust on articulated objects, see appendix \ref{sec:supp_robustness}.

\vspace{-1mm}
\subsection{Tracking speed}
\vspace{-1mm}
Aside from the state-of-the-art performance, our method also has the highest tracking speed among all others, as summarized in Table \ref{table:fps}. All methods are tested on the same device. NOCS and ANCSH are slow due to RANSAC and optimizations, which we don't need. 6-PACK's actual speed is slower than what they claim in their paper ($>$10 FPS) since the network in their officially released code \cite{j96w} is forwarded 27 times at a grid of potential object centers at each frame to achieve their reported performance.

\begin{center}
\begin{table}[t]
\centering
\begin{tabular}{|c|c|c|c|c|}
\hline
Method  & NOCS  & 6-PACK & ANCSH & Ours \\ \hline
FPS & 4.05 & 3.53 & 0.80 & \textbf{12.66} \\ \hline
\end{tabular}
\vspace{1mm}
\caption{\textbf{Tracking speed in FPS}}
\label{table:fps}
\vspace{-5mm}
\end{table} 
\end{center}

\vspace{-10mm}
\subsection{Discussions}\label{sec:discussion}
\vspace{-1mm}
\noindent\textbf{Tracking Scale} Although the actual scale of the object is constant during tracking, we still track the scale in our framework to deal with inaccurate initial scale. Compared to fixing the scale as the noisy initial scale throughout the tracking, our scale tracking scheme decreases the average scale error from 1.09\% to 0.30\% and increases mIoU from 71.70\% to 74.00\% on articulated objects; and improve mIoU from 73.43\% to 76.42\% on rigid objects (excluding camera where both schemes fail).

\noindent\textbf{Leveraging Kinematic Chain Constraints} For articulated objects, our method focuses on per-part tracking without explicitly leveraging joint constraints at test time.
Prior works leverage these constraints in instance-level tracking \cite{BMVC2015_181,desingh2019factored} and category-level tracking \cite{li2020categorylevel}. \cite{BMVC2015_181} and \cite{desingh2019factored} assume perfect knowledge of joint parameters and treat them as a hard constraint. In the category-level setting, however, joint parameters are unknown and difficult to predict due to occlusions, especially for pivot point predictions. 
Empirically, we have tried predicting them and achieved accuracy similar to state-of-the-art \cite{li2020categorylevel}, \textit{e.g.} 1.1cm  error for laptop pivot points. However, enforcing these inaccurate constraints harmed our performance, leading to an 80\% increase in translation error.
ANCSH \cite{li2020categorylevel} offers an alternative where only estimated joint axis orientations are used as soft constraints for rotation predictions at the cost of lower speed. Note that without leveraging the constraints, our method already significantly outperforms ANCSH \cite{li2020categorylevel}. Without sacrificing speed, we examine the usage of ground truth joint axis orientations as hard constraints but only gain little improvement as shown in Table \ref{table:articulated} ({Ours + Rot. Proj.}). We leave this direction to future works.

\noindent\textbf{Limitations and Failure Cases}\label{sec:failure_case}
Most of our failure cases come from large sensor noise in real depth images. In extreme cases, \textit{e.g.} on real cameras from NOCS-REAL275 which are either too reflective or too dark, our CoordinateNet fails to produce reasonable segmentation and the whole pipeline collapses (see appendix \ref{sec:per_category_rigid}). In milder cases, domain gap resulting from sensor noise also degrades our performance. Specific domain adaptation techniques may be needed to deal with this issue, which are beyond the scope of this paper.
\vspace{-2mm}

\section{Conclusion}
\vspace{-1mm}
In this paper, for the first time, we tackle the problem of category-level 9DoF pose tracking for both rigid and articulated objects. To achieve this goal, we propose an end-to-end differentiable pose tracking framework consisting of three modules: pose canonicalization, RotationNet, and CoordinateNet. Our algorithm achieves state-of-the-art performance on both category-level rigid and articulated pose benchmarks and runs comparably fast for evaluation.

\noindent \textbf{Acknowledgement:}
This research is supported by a grant from the SAIL-Toyota Center for AI Research, a grant from the Samsung GRO program, NSF grant IIS-1763268, a Vannevar Bush Faculty fellowship, the support of the Stanford UGVR program, and gifts from Kwai and Qualcomm. Toyota Research Institute ("TRI") provided funds to assist the authors with their research but this article solely reflects the opinions and conclusions of its authors and not TRI or any other Toyota entity.

{\small
\bibliographystyle{ieee_fullname}
\bibliography{egbib}
}

\begin{appendix}
\clearpage
\section{Coordinate-based Method Review}\label{sec:nocs}
\vspace{-1mm}
\paragraph{Normalized Object Coordinate Space (NOCS)}
NOCS is a category-level canonical reference frame defined within a unit 3D cube introduced in \cite{wang2019normalized}. The objects from the same category in NOCS are consistently aligned to a category-level canonical orientation. These objects are further zero-centered and uniformly scaled so that their tight bounding boxes are centered at the origin of NOCS with a diagonal length of 1. 

Mathematically, for an object point cloud $X\in \mathbb{R}^{N\times 3}$, its corresponding point-wise normalized object coordinates are denoted as $Y\in\mathbb{R}^{N\times 3}$. The transformation between $Y$ in the NOCS frame and $X$ in the camera frame is a 7D similarity transformation $\mathcal{T}^{(j)} = \{s^{(j)}, R^{(j)}, T^{(j)}\}$, which satisfies $X = sRY + T$. This 7D transformation defines the category-level 6D pose and 1D uniform scale of rigid objects.

Given the input object point cloud $X$, Wang \textit{et. al.} \cite{wang2019normalized} trained a deep neural network to directly regress $Y$. The 7D similarity transformation can then be computed with the 3D-3D point correspondence established between $X$ and $Y$ using the Umeyama algorithm \cite{umeyama1991least} along with RANSAC. 
Knowing the 1D scale $s$, the actual object size $d \in \mathbb{R}^3$ can be estimated as $d = s\times(|x|_{max}, |y|_{max}, |z|_{max})$.

\noindent\textbf{Normalized Part Coordinate Space (NPCS)} 
Li \textit{et. al.}\cite{li2020categorylevel} extended the definition of NOCS to rigid parts in articulated objects, and proposed a part-level canonical reference frame, namely NPCS. Similar to NOCS, each individual part has canonical orientation, zero translation, and normalized scale in its NPCS. Leveraging per-point NPCS estimations, per-part 9DoF poses can be estimated in the same way as NOCS.

\vspace{-2mm}
\section{Proof of the pose canonicalization observation}\label{sec:proof}
\vspace{-1mm}
\begin{proof}
Denote the corresponding points of $C_{t+1}^{(j)}$ in $Z_{t+1}^{(j)}$ as $\hat{C}_{t+1}^{(j)}$, then $C_{t+1}^{(j)} = s_{t + 1}^{(j)}R_{t + 1}^{(j)}Y_{t + 1}^{(j)} + T_{t + 1}^{(j)}$, $\hat{C}_{t + 1}^{(j)}= \hat{s}_t^{(j)} \hat{R}_t^{(j)}Y_{t + 1}^{(j)} + \hat{T}_{t}^{(j)}$.
Given that $\hat{C}_{t+1}^{(j)} =  (R_t^{(j)})^{-1}(C_{t+1}^{(j)} - T_t^{(j)})/s_t^{(j)}$, we can obtain $\hat{s}_t^{(j)} = (s_t^{(j)})^{-1}s_{t+1} \approx 1$, $\hat{R}_t^{(j)} = (R_t^{(j)})^{-1}R_{t + 1}^{(j)} \approx I$, $\hat{T}_{t}^{(j)} = (s_t^{(j)})^{-1}(R_t^{(j)})^{-1}(T_{t + 1}^{(j)} - T_t^{(j)})\approx 0$ , as a natural result of temporal continuity.
\end{proof}

\vspace{-3mm}
\section{Principal Component Analysis of Residual Rotations}\label{sec:pca}
\vspace{-1mm}
To probe the structure of RotationNet's output space, i.e. the residual rotations in pose-canonicalized point clouds, we perform Principal Component Analysis on $N = 10^6$ random rotations with rotation angle $\le \theta \in \{5^{\circ}, 10^{\circ}, 15^{\circ}, 30^{\circ}, 60^{\circ} 180^{\circ}(\text{unconstrained})\}$, in the form of the 6D continuous representation. To obtain random rotations, we first uniformly sample rotations in $SO(3)$, then filter them according to their rotation angles. Table \ref{table:pca} shows the ratio of variance explained by the first three principal components, denoted $r_{1}, r_{2}, r_{3}$. In our case where residual rotations are small, the linear space spanned by the first three components captures most of the variance of the output space.This is simply because: 1) the rotation space has 3 degree of freedom; 2) when the rotation is limited to a small range around identity, this space is almost linear. 

Thanks to this linearity, we can use RotationNet to directly regress the small residual rotation with high accuracy.

\begin{table}[h]
\begin{tabular}{|c|ccc|c|}
\hline
$\theta$      & $r_1$ & $r_2$ & $r_3$ & $r_1+r_2+r_3$ \\ \hline
$5^{\circ}$   & 49.81\%       & 25.20\%       & 24.94\%       & 99.95\%             \\
$10^{\circ}$  & 49.85\%       & 25.10\%       & 24.86\%       & 99.81\%             \\
$15^{\circ}$  & 49.74\%       & 24.90\%       & 24.83\%       & 99.46\%             \\
$30^{\circ}$  & 48.91\%       & 24.73\%       & 24.69\%       & 98.33\%             \\
$60^{\circ}$  & 45.62\%       & 23.86\%       & 23.77\%       & 93.25\%             \\
$180^{\circ}$ & 16.73\%       & 16.69\%       & 16.68\%       & 50.10\%             \\ \hline
\end{tabular}
\vspace{1mm}
\caption{\textbf{Ratio of variance explained by the first three principal components.}}
\label{table:pca}
\vspace{-4mm}
\end{table}

\vspace{-3mm}
\section{Per-part Rotation, Scale and Translation Computation}
\vspace{-1mm}
\subsection{Euclidean Mean of Rotations}
\vspace{-1mm}
For averaging over rotations, we adopt the euclidean mean \cite{moakher2002means} of the multiple rotation predictions, which converts the 6D rotation representation back to matrix format, takes the mean matrix, and then project back to $SO(3)$. Taking a binary segmentation mask $m_{t+1}^{(j)}$, our final prediction is given by $\hat{R}_{t + 1}^{(j)}\ = \text{EuclideanMean}(\{\hat{R}_{i,t + 1}^{(j)} | i\in m_{t+1}^{(j)}\}_i)$.

\vspace{-1mm}
\subsection{Rotation Supervision for Symmetric Objects}\label{sec:sym_rotation}
\vspace{-2mm}
Unseen instances from symmetric object categories, like bowls or bottles contain a rotation ambiguity around their symmetric axis $\hat{q}$ \cite{wang2019normalized}. 
Due to this rotation ambiguity, only two degrees of freedom in the rotation are unique and should be supervised. We propose to regress the 3D end-point position $p_{\hat{R}}$ of its unit rotation axis $\hat{q}$. Similar to \cite{zhou2019continuity}, the redundancy in the representation renders a continuous and regression-friendly rotation representation for symmetric objects. On the other hand, articulated objects rarely have rotational ambiguities for their rigid parts, since their kinematic structures usually help to disambiguate.
Therefore, we only use the symmetric rotation representation for the bowl, bottle, and can categories in the NOCS-REAL275 dataset \cite{wang2019normalized}.

\vspace{-2mm}
\subsection{Coordinate Supervision for Symmetric Objects}\label{sec:sym_coord}
\vspace{-2mm}
For a symmetric object, \textit{e.g.}, a bowl, its normalized coordinates contain ambiguities: one can freely rotate them together along its symmetric axis. Note that point pairwise distances are invariant under the rotation as are their $y$ and $\sqrt{x^2 + z^2}$ values ($y$ is the symmetric axis). To supervise coordinate predictions for symmetric objects, we propose to jointly enforce an L2 loss on the pairwise distance matrix and a symmetric coordinate loss $\sqrt{|x^2 + z^2 - \hat x^2 - \hat z^2| + (y - \hat y^2)}$ on the normalized coordinates.

\begin{table*}[t!]
\begin{tabular}{|c|c|c|c|c|c|c|c|c|c|}
\hline
\multirow{4}{*}{Category} & \multicolumn{4}{c|}{\multirow{2}{*}{Part definitions}}                                                & \multicolumn{2}{c|}{\multirow{2}{*}{Data statistics}}                     & \multicolumn{3}{c|}{\multirow{2}{*}{\shortstack{Training Pose Perturbation \\ Distribution $\mathcal{N}(0, \sigma)$}}}                       \\
                          & \multicolumn{4}{c|}{}                                                                                 & \multicolumn{2}{c|}{}                                                     & \multicolumn{3}{c|}{}                                                                                                   \\ \cline{2-10} 
                          & \multirow{2}{*}{Part 0} & \multirow{2}{*}{Part 1} & \multirow{2}{*}{Part 2} & \multirow{2}{*}{Part 3} & \multirow{2}{*}{Train/Test} & \multirow{2}{*}{\shortstack{Average joint \\ state change}} & \multirow{2}{*}{$\sigma_{scale}$} & \multirow{2}{*}{$\sigma_{rot}$($^{\circ}$)} & \multirow{2}{*}{$\sigma_{trans}\text{(cm)}$} \\
                          &                         &                         &                         &                         &                             &                                             &                                   &                                             &                                       \\ \hline
glasses                   & right temple            & left temple             & base                    & -                       & 47/8                        & 19.19$^{\circ}$                             & 0.02                              & 5                                           & 2                                     \\
scissors                  & right half              & left half               & -                       & -                       & 33/3                        & 34.32$^{\circ}$                             & 0.01                              & 3                                           & 1                                     \\
laptop                    & base                    & display                 & -                       & -                       & 49/6                        & 26.13$^{\circ}$                             & 0.015                             & 3                                           & 2                                     \\
drawers                   & lowest                  & middle                  & top                     & base                    & 28/2                        & 3.72cm                                      & 0.02                              & 3                                           & 2                                     \\ \hline
\end{tabular}
\vspace{1mm}
\caption{\textbf{Statistics of our synthetic articulated object dataset.}}
\label{table:sapien_dataset}
\vspace{-4mm}
\end{table*}

\vspace{-2mm}
\subsection{Scale and Translation Computation}\label{sec:compute_st}
\vspace{-2mm}
For part $j$, given segmentation mask prediction $\tilde{m}_{t + 1}^{(j)}$ and normalized coordinate predictions from the CoordinateNet, we can obtain input points from part $j$, namely $\tilde{C}_{t + 1}^{(j)} = \{X_{i, t + 1} | i \in \tilde{m}_{t + 1}^{(j)}\}$ and their corresponding normalized coordinate predictions $\tilde{Y}_{t + 1}^{(j)}$.

Based on RotationNet predictions, we can compute absolute per-part rotation prediction $\tilde{R}_{t + 1}^{(j)} = \tilde{R}_{t}^{(j)}\hat{R}_{t + 1}^{(j)}$. 

For asymmetrical objects, let $\tilde{W}_{t + 1}^{(j)} = \tilde{R}_{t + 1}^{(j)} \tilde{Y}_{t + 1}^{(j)}$, $\tilde{W}_{t + 1}^{(j)}$ and $\tilde{C}_{t + 1}^{(j)}$ only differs by a scaling and a translation, namely
\vspace{-2mm}
$$\tilde{C}_{t + 1}^{(j)} = \tilde{s}_{t + 1}^{(j)}\tilde{W}_{t + 1}^{(j)} + \tilde{T}_{t + 1}^{(j)}$$
\vspace{-2mm}
We compute the scale and translation of part $j$ as follows:

$$\tilde{s}_{t + 1}^{(j)} = \sum_{i} \tilde{W}_{i, t + 1}^{(j)\top}\tilde{C}_{i, t + 1}^{(j)} / \sum_{i} \tilde{W}_{i, t + 1}^{(j)\top}\tilde{W}_{i, t + 1}^{(j)}$$
\vspace{-1mm}
$$\tilde{T}_{t + 1}^{(j)} = \operatorname{avg}_{i}(\tilde{C}_{i, t + 1}^{(j)} - \tilde{s}_{t + 1}^{(j)}\tilde{W}_{i, t + 1}^{(j)})$$

For symmetrical objects, Let $\tilde{U}_{t + 1}^{(j)} = \tilde{R}_{t + 1}^{(j)\top}\tilde{C}_{t + 1}^{(j)}$. Besides a scaling and a translation, $\tilde{Y}_{t + 1}^{(j)}$ and $\tilde{U}_{t + 1}^{(j)}$ may further differ by a rotation $R(l, \theta)$ around the axis of symmetry $l$, namely 

\vspace{-4mm}
$$\tilde{U}_{t + 1}^{(j)} = \tilde{s}_{t + 1}^{(j)}R(l, \theta)\tilde{Y}_{t + 1}^{(j)} + \tilde{R}_{t + 1}^{(j)\top}\tilde{T}_{t + 1}^{(j)}$$
\vspace{-4mm}

This is because for ground-truth normalized coordinates $Y_{t + 1}^{(j)*}$, $R(l, \theta)Y_{t + 1}^{(j)*}$ will also be a correct set of predictions due to the object symmetry.

To simplify the problem, we assume $l$ overlaps with the $y$-axis, then $R(l, \theta)$ becomes a 2D rotation in $xz$-plane. 

We propose to take the $xz$-plane projection of everything and use the 2D version of Umeyama algorithm \cite{umeyama1991least} to compute $R(l, \theta)$. Then we have:
\vspace{-2mm}
$$R(l, \theta)^{\top}\tilde{U}_{t + 1}^{(j)} = \tilde{s}_{t + 1}^{(j)}\tilde{Y}_{t + 1}^{(j)} + (\tilde{R}_{t + 1}^{(j)}R(l, \theta))^{\top}\tilde{T}_{t + 1}^{(j)}$$

This is the same case as asymmetrical objects, we can compute $\tilde{s}_{t + 1}^{(j)}$ and $\tilde{T}_{t + 1}^{(j)}$ similarly.

\section{Implementation Details}

\subsection{Network Input}
For synthetic articulated data, the input to our network is a partial depth point cloud projected from a single-view depth image, downsampled to $N=4096$ points using FPS. For NOCS-REAL275 data, our pipeline crops a ball centered at estimated position of the object center, with a radius $1.2$ times the object's estimated radius. Scene points within the ball are then downsampled to $N=4096$ points.

\subsection{Training Details}
Our network is implemented using PyTorch and optimized by the Adam optimizer, with a learning rate starting at ${10}^{-3}$ and decay by half every $20$ epochs. It takes around 20 and 100 epochs for our model to converge on the rigid object dataset and the articulated object dataset, respectively. We have made our code public at \href{https://github.com/halfsummer11/CAPTRA}{https://github.com/halfsummer11/CAPTRA}

\subsection{Network Architecture}

Both CoordinateNet and RotationNet use PointNet++ \cite{qi2017pointnet++} MSG segmentation network as their backbone. The detailted architecture is as follows:
\begin{equation*}
\begin{aligned}
&\text{Backbone:} \\
&~~\text{SA}(\texttt{num\_points}=512, \texttt{radius}=[0.05, 0.1, 0.2], \\
&~~\qquad \texttt{mlps=}[[32, 32, 64], [64, 64, 128], [64, 96, 128]]) \rightarrow \\
&~~\text{SA}(\texttt{num\_points}=128, \texttt{radius}=[0.2, 0.4], \\ 
&~~\qquad \texttt{mlps=}[[128, 128, 256], [128, 196, 256]]) \rightarrow \\
&~~\text{GlobalSA}(\texttt{mlp=}[256, 512, 1024])\rightarrow \\
\end{aligned}
\end{equation*}
\begin{equation*}
\begin{aligned}
&~~\text{FP}(\texttt{mlp=}[256, 256]) \rightarrow \\
&~~\text{FP}(\texttt{mlp=}[256, 128]) \rightarrow \\
&~~\text{FP}(\texttt{mlp=}[128, 128]) \\
&\text{CoordinateNet:}  \\
&~~\text{Backbone} \rightarrow \\
&~~~~~~\text{Coordinate Head:}\text{MLP}([128, 3P]) \rightarrow \text{Sigmoid}() \\
&~~~~~~\text{Segmentation Head:} \text{FC}([P + 1])
\rightarrow \text{Softmax}() \\
&\text{RotationNet:} \\
&~~\text{Backbone}  \rightarrow \text{MLP}([512, 512, 256, 6P])  \\
\end{aligned}
\end{equation*}
We use LeakyReLU and group normalization for each FC layer in set abstraction (SA) and feature propagation (FP) layers.

\section{SAPIEN Articulated Objects Data Generation and Statistics}\label{sec:sapien_dataset}

We render our synthetic articulated object pose tracking dataset using SAPIEN \cite{xiang2020sapien}. In Table \ref{table:sapien_dataset}, we summarize for each object category 1) the part definitions; 2) the train-test split; 3) the average joint state change over all test sequences (each consisting of 100 frames); and 4) the variance of Gaussian noise distributions from which we sample input pose perturbations during training. \textbf{Note that both the global pose and the joint states of the articulated objects are changing in the trajectories.} 

We use different amount of noise for different object categories depending on the difficulty of pose estimation, e.g. the poses of thin glasses temples are difficult to predict, therefore we train the model with a larger perturbation to handle larger prediction error during tracking.

\section{Experiment Details of Real Drawers under Robot-Object Interaction}\label{sec:real_art_detail}

\vspace{-1mm}
\begin{figure}[h]
\begin{center}
   \includegraphics[width=0.8\linewidth]{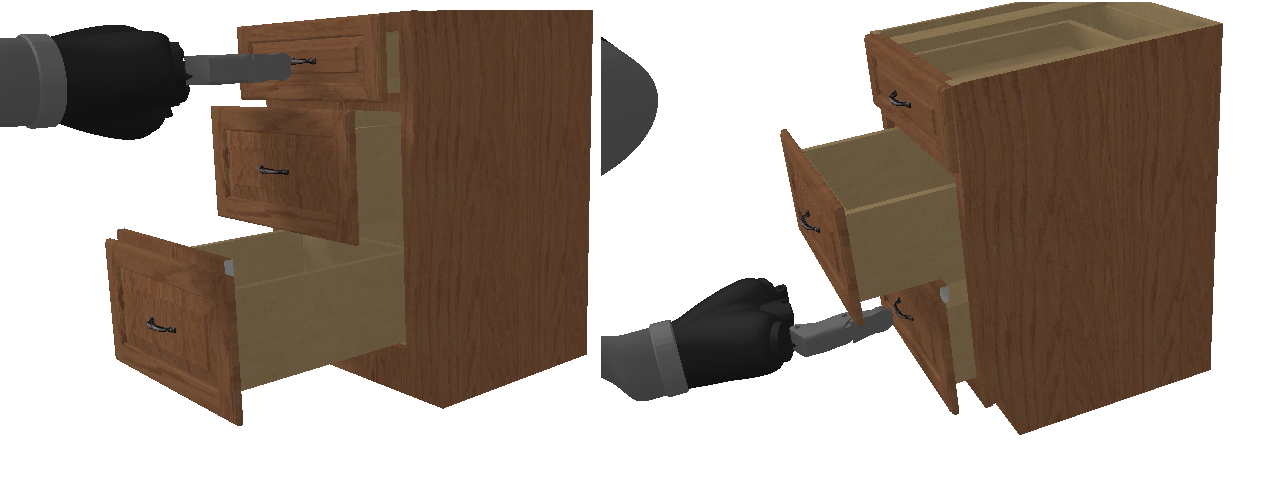}
\end{center}
    \vspace{-3mm}
   \caption{\textbf{Synthetic training data for real drawers tracking.} We use SAPIEN environment to render depth sequences where a Kinova Jaco2 robotic arm manipulates one drawer. Here color images are for visualization only.}
   \vspace{-3mm}
   \label{fig:real_drawers_train}
\end{figure}

To mimic the interaction scenario, we simulate a Kinova Jaco2 robotic arm in the SAPIEN \cite{xiang2020sapien} environment, make it push synthetic drawers from the SAPIEN dataset, and render depth images. Figure \ref{fig:real_drawers_train} shows two examples of our simulated data. We train our model only on these synthetic depth images following the same protocol as in Section 5.3.

\begin{center}
\begin{table}[t!]
\centering

\setlength{\tabcolsep}{3pt}
\resizebox{0.5\textwidth}{!}{
\begin{minipage}{0.64\textwidth}
\begin{tabular}{|c|c|c|c|c|c|c|c|c|}
\hline

\multicolumn{2}{|c|}{\multirow{2}{*}{Method}}          & \multirow{2}{*}{NOCS}  & \multirow{2}{*}{CASS}  & \multirow{2}{*}{CPS++} & \multirow{2}{*}{\shortstack{Oracle\\ICP}} & \multirow{2}{*}{6PACK}    & \multirow{2}{*}{6PACK}    & \multirow{2}{*}{Ours} \\ 
\multicolumn{2}{|c|}{}  &   &   &   &   &   &   & \\ \hline
\multicolumn{2}{|c|}{Input}           & RGBD  & RGBD  & RGB & Depth & RGBD & RGBD & Depth \\\hline 
\multicolumn{2}{|c|}{Setting}         & \multicolumn{3}{|c|}{Single frame} & \multicolumn{4}{|c|}{Tracking} \\ \hline

\multicolumn{2}{|c|}{Initialization}           & N/A   & N/A   &  N/A  & GT. & GT.  & Pert.  & Pert. \\ \hline
 \multirow{4}{*}{bottle}  & 5$^{\circ}$5cm $\uparrow$ & 5.50  & 11.49 & 2.90  & 0.28  & 14.11 & 17.48 & \textbf{{79.46}} \\
 & mIoU$\uparrow$            & 33.73 & 34.72 & 27.91 & 10.72 & 59.77 & 49.98 & \textbf{{72.11}} \\
 & $R_{err}\downarrow$       & 25.60 & 18.39 & 14.81 & 44.03 & 21.45 & 12.83 & \textbf{{3.29}}  \\
 & $T_{err}\downarrow$       & 14.40 & 26.66 & 32.67 & 8.28  & 3.36  & 4.97  & \textbf{{2.60}}  \\ \hline
 
 \multirow{4}{*}{bowl} & 5$^{\circ}$5cm $\uparrow$ & 62.20         & 33.50 & 5.61  & 0.45  & 40.46 & 34.30 & \textbf{{79.20}} \\
 & mIoU$\uparrow$            & 78.78         & 62.36 & 32.07 & 11.54 & 56.29 & 56.15 & \textbf{{79.64}} \\
 & $R_{err}\downarrow$       & 4.70          & 5.98  & 12.39 & 30.31 & 5.83  & 6.78  & \textbf{{3.50}}  \\
 & $T_{err}\downarrow$       & \textbf{{1.20}} & 4.76  & 19.97 & 6.65  & 1.64  & 1.67  & {1.43}  \\ \hline
 
 \multirow{4}{*}{can} & 5$^{\circ}$5cm $\uparrow$ & 7.10          & 22.24 & 3.22  & 0.49  & 28.07 & 21.51 & \textbf{{64.70}} \\
 & mIoU$\uparrow$            & 49.56         & 59.43 & 33.20 & 10.50 & 50.32 & 49.48 & \textbf{{62.47}} \\
 & $R_{err}\downarrow$       & 16.90         & 12.08 & 13.99 & 43.85 & 11.66 & 16.58 & \textbf{{3.43}}  \\
 & $T_{err}\downarrow$       & \textbf{{4.00}} & 9.08  & 19.75 & 8.48  & 5.03  & 5.82  & {5.69}  \\ \hline
 
 \multirow{4}{*}{camera} & 5$^{\circ}$5cm $\uparrow$ & 0.60          & \textbf{{12.73}} & 0.29  & 0.60  & 6.89  & 0.97  & {0.41}  \\
 & mIoU$\uparrow$            & 58.13         & \textbf{{60.84}} & 36.18 & 19.62 & 52.10 & 51.55 & {2.50}  \\
 & $R_{err}\downarrow$       & 33.80         & \textbf{{14.70}} & 30.22 & 36.09 & 49.96 & 57.65 & {17.82} \\
 & $T_{err}\downarrow$       & \textbf{{3.10}} & 7.29           & 16.12 & 7.23  & 6.06  & 5.65  & {35.53} \\ \hline
 
 \multirow{4}{*}{laptop}  & 5$^{\circ}$5cm $\uparrow$ & 25.50         & 82.81 & 0.51  & 1.60  & 64.09 & 36.31 & \textbf{{94.03}} \\
 & mIoU$\uparrow$            & 52.59         & 63.98 & 19.58 & 22.11 & 49.76 & 49.79 & \textbf{{87.20}} \\
 & $R_{err}\downarrow$       & 8.60          & 5.89  & 30.85 & 14.39 & 5.03  & 6.12  & \textbf{{2.24}}  \\
 & $T_{err}\downarrow$       & {2.40} & 3.89  & 13.47 & 8.41  & 2.57  & 2.44  & \textbf{{1.48}}  \\ \hline
 
 \multirow{4}{*}{mug} & 5$^{\circ}$5cm $\uparrow$ & 0.90          & {13.85} & 0.90  & 0.48  & 19.90 & 22.23 & \textbf{{55.17}} \\
 & mIoU$\uparrow$            & 58.08         & {54.56} & 31.15 & 13.63 & 64.26 & 64.54 & \textbf{{80.70}} \\
 & $R_{err}\downarrow$       & 31.50         & {27.97} & 49.65 & 73.02 & 22.06 & 17.99 & \textbf{{5.36}}  \\
 & $T_{err}\downarrow$       & {4.00} & 20.76          & 27.73 & 7.21  & 1.19  & 1.17  & \textbf{{0.79}}  \\ \hline
 
 \multirow{4}{*}{all} & 5$^{\circ}$5cm $\uparrow$ & 16.97         & {29.44} & 2.24  & 0.65  & 28.92         & 22.13 & \textbf{{62.16}} \\
 & mIoU$\uparrow$            & 55.15         & {55.98} & 30.02 & 14.69 & 55.42         & 53.58 & \textbf{{64.10}} \\
 & $R_{err}\downarrow$       & 20.18         & {14.17} & 25.32 & 40.28 & 19.33         & 19.66 & \textbf{{5.94}}  \\
 & $T_{err}\downarrow$       & {4.85} & 12.07          & 21.62 & 7.71  & \textbf{{3.31}} & 3.62  & {7.92}  \\ \hline
 
 \multirow{4}{*}{\shortstack{all\\w/o\\ cam.}}  & 5$^{\circ}$5cm $\uparrow$ & 20.24         & {32.78} & 2.63  & 0.66  & 33.33         & 26.37 & \textbf{{74.51}} \\
 & mIoU$\uparrow$            & 54.55         & {55.01} & 28.78 & 13.70 & 56.08         & 53.99 & \textbf{{76.42}} \\
 & $R_{err}\downarrow$       & 17.46         & {14.06} & 24.34 & 41.12 & 13.20         & 12.06 & \textbf{{3.56}}  \\
 & $T_{err}\downarrow$       & {5.20} & 13.03          & 22.72 & 7.81  & {2.76} & 3.21  & \textbf{{2.40}}  \\ \hline
\end{tabular}
\end{minipage}
}
\vspace{1mm}
\caption{\textbf{Per-category results of category-level rigid object pose tracking on NOCS-REAL275}}.
\label{table:rigid_per_category}
\vspace{-5mm}

\end{table} 
\end{center}

\begin{center}
\begin{table}[h]
\begin{tabular}{|c|c|c|c|}
\hline
Mask Source              & CoordNet(Depth) & NOCS(RGB) & GT    \\ \hline
5$^{\circ}$5cm$\uparrow$ & 0.41            & 9.05      & 20.09 \\
mIoU$\uparrow$           & 2.50            & 33.00     & 46.35 \\
$R_{err}\downarrow$      & 17.82           & 20.75     & 10.89 \\
$T_{err}\downarrow$      & 35.53           & 13.09     & 3.67  \\ \hline
\end{tabular}
\vspace{1mm}
\caption{\textbf{Results on the camera category from NOCS-REAL275 with different segmentation mask sources.}}
\label{table:camera_seg}
\vspace{-5mm}
\end{table}
\end{center}

\begin{center}
\begin{table*}[t!]
\centering
\setlength{\tabcolsep}{3pt}
\resizebox{1.0\textwidth}{!}{
\begin{tabular}{|c|c|c|c||c||c|c|c|c|}
\hline

\multicolumn{2}{|c|}{\multirow{2}{*}{Method}}  & \multirow{2}{*}{ANCSH}  & \multirow{2}{*}{\shortstack{Oracle \\ ICP}}  & \multirow{2}{*}{Ours} & \multirow{2}{*}{$C$-sRT} & \multirow{2}{*}{$C$-CrdNet} & \multirow{2}{*}{\shortstack{$C$-Crd.+\\DSAC++ }} & \multirow{2}{*}{\shortstack{Ours w/o \\$L_c, L_s, L_t$}}  \\ 

\multicolumn{2}{|c|}{}       &               &         &       &       &      &   &         \\ \hline

 \multirow{5}{*}{glasses} 
 & 5$^{\circ}$5cm$\uparrow$ & 72.6, 75.8, 81.9 & 46.9, 46.1, 78.4   & \textbf{97.7}, 95.3, \textbf{99.6} & 27.1, 22.6, 25.3    & 91.6, 89.2, 91.5          & 81.9, 84.0, 92.4 & 97.4, \textbf{96.0}, 99.2 \\
 & mIoU$\uparrow$           & 73.7, 74.3, 47.7 & 65.8, 67.2, 56.0   & \textbf{81.8, 81.4, 57.2} & 12.6, 11.5, 1.7     & 81.2, 80.8, 56.7          & 67.7, 71.2, 41.5 & 80.8, 80.8, 55.0 \\
 & $R_{err}\downarrow$      & 4.17, 3.86, 3.58 & 11.00, 10.22, 4.66 & \textbf{1.72, 1.93, 1.22} & 5.80, 5.86, 2.98    & 2.78, 3.06, 1.90          & 3.43, 3.17, 2.00 & 1.87, 2.14, 1.47 \\
 & $T_{err}\downarrow$      & 0.47, 0.50, 0.23 & 2.10, 2.82, 1.98   & 0.27, 0.26, \textbf{0.14}          & 11.43, 12.56, 12.67 & \textbf{0.25, 0.24, 0.14} & 0.55, 0.39, 0.22 & 0.27, 0.33, 0.17 \\
 & $\theta_{err}\downarrow$ & 1.40, 1.43       & 5.25, 4.25         & 0.94, 0.97                & 3.26, 3.64          & \textbf{0.65}, 0.74       & 0.73, \textbf{0.65 }      & 0.93, 1.14       \\ \hline

 \multirow{5}{*}{scissors} 
& 5$^{\circ}$5cm$\uparrow$ & 98.7, 98.8 & 25.7, 28.3   & 99.0, 99.4    & 3.1, 2.7     & 96.6, 98.7          & \textbf{99.5, 99.9} & 98.4, 99.4          \\
 & mIoU$\uparrow$           & 64.0, 64.4 & 19.9, 26.8   & 65.6, 71.9    & 1.1, 1.6     & 64.9, 72.5          & \textbf{65.7}, 71.4 & 63.0, \textbf{72.2}          \\
 & $R_{err}\downarrow$      & 1.82, 1.77 & 19.85, 17.30 & 1.60, \textbf{1.17}    & 55.17, 59.08 & 2.25, 1.88          & 1.56, 1.77          & \textbf{1.48}, 1.23 \\
 & $T_{err}\downarrow$      & 0.16, 0.21 & 7.80, 4.82   & 0.12, 0.14    & 7.63, 7.62   & \textbf{0.10, 0.12} & 0.13, 0.14          & 0.14, 0.15          \\
 & $\theta_{err}\downarrow$ & 1.96       & 13.14        & \textbf{1.85} & 11.11        & 1.93                & 1.96                & 1.89                \\ \hline
 
 \multirow{5}{*}{laptop} 
 & 5$^{\circ}$5cm$\uparrow$ & \textbf{97.5, 99.1} & 81.4, 92.4 & 97.1, 97.2          & 38.8, 57.3 & 96.1, 98.3 & 96.5, 98.4 & 96.6, 95.9 \\
 & mIoU$\uparrow$           & 70.3, 50.6          & 52.5, \textbf{62.7} & \textbf{76.2}, 53.5 & 45.9, 40.4 & 74.3, 54.0 & 47.5, 42.7 & 73.2, 47.8 \\
 & $R_{err}\downarrow$      & 1.72, 1.08          & 6.85, 1.70 & \textbf{0.62, 1.22} & 4.86, 2.61 & 3.02, 1.92 & 2.14, 1.49 & 1.18, 1.31 \\
 & $T_{err}\downarrow$      & 0.58, 0.49          & 2.00, 0.90 & \textbf{0.32, 0.35} & 9.63, 6.18 & 0.58, 0.52 & 1.31, 1.00 & 0.43, 0.42 \\
 & $\theta_{err}\downarrow$ & 1.48                & 3.74       & \textbf{1.33}       & 3.69       & 1.94       & 2.17       & 1.35       \\ \hline
  
 \multirow{5}{*}{drawers} 
 & 5$^{\circ}$5cm$\uparrow$ & 94.3, 93.5, 98.1, 99.6 & 65.8, 79.7, 79.9, 96.4 & \textbf{99.6, 99.6, 99.6, 99.7} & 6.7, 11.2, 14.1, 11.3      & 92.2, 91.7, 97.2, 97.4 & 97.2, 96.7, 97.8, 97.5 & 97.4, 97.3, 98.0, 98.5 \\
 & mIoU$\uparrow$           & 80.7, 83.3, 84.4, 91.1 & 73.8, 80.8, 82.3, 93.3 & \textbf{85.1, 86.4, 89.8, 94.2} & 26.2, 30.3, 30.9, 41.2     & 83.5, 84.7, 88.8, 93.0 & 84.2, 85.2, 88.2, 88.2 & 84.9, 86.3, 88.9, 92.0 \\
 & $R_{err}\downarrow$      & 2.11, 2.21, 1.67, 0.69 & 8.45, 5.40, 2.69, 0.80 & \textbf{0.18, 0.18, 0.19, 0.23} & 22.07, 15.82, 16.20, 23.39 & 2.22, 2.20, 1.23, 0.63 & 1.18, 1.21, 0.83, 0.70 & 0.55, 0.65, 0.44, 0.51 \\
 & $T_{err}\downarrow$      & 1.15, 0.85, 0.68, 0.51 & 3.33, 2.51, 1.48, 1.07 & \textbf{0.59, 0.60, 0.38, 0.29} & 22.99, 17.56, 13.07, 18.77 & 0.91, 0.93, 0.46, 0.57 & 0.70, 0.62, 0.40, 0.66 & 0.74, 0.73, 0.51, 0.36 \\
 & $d_{err}\downarrow$ & 0.72, 0.62, 0.58       & 1.33, 1.00, 0.82       & \textbf{0.37, 0.36, 0.28}       & 5.31, 6.91, 10.48          & 0.75, 0.83, 0.68       & 0.46, 0.65, 0.58       & 0.39, 0.37, 0.32       \\ \hline
 
\end{tabular}
}
\vspace{1mm}
\caption{\textbf{Per-part, per-category results of category-level articulated object pose tracking on held-out instances from SAPIEN.}}
\label{table:articulated_full}
\end{table*} 
\end{center}
\vspace{-15mm}

\section{Discussions on not Using RANSAC}
Most coordinate-based pose estimation approaches heavily rely on RANSAC during pose fitting, because rotations estimation done by orthogonal Procrustes is very sensitive to outliers. In our pipeline, the pose canonicalization significantly simplifies the rotation regression and reduce the noise in the coordinate prediction, thus freeing us from the need to use RANSAC.
Our experiment shows that incorporating RANSAC to our scale and translation computation only bring very little improvements, i.e., increasing 5$^{\circ}$5cm accuracy and mIoU by 0.86\% and 1.85\% respectively for rigid objects from NOCS-REAL275, 0.06\% and 0.06\% respectively for articulated objects from SAPIEN. 
In contrast, when we remove RANSAC, {C-CoordinateNet} drops 3.22\% on 5$^{\circ}$5cm accuracy and 4.37\% on mIoU, due to the aforementioned rotation sensititity.

\section{Additional Results}

\subsection{Per-Category Results for Rigid Objects}\label{sec:per_category_rigid}
Table \ref{table:rigid_per_category} summarizes the per-category results for rigid object pose tracking on NOCS-REAL275. Our method only fails on the camera category and outperforms the previous state-of-the-arts under most metrics on all other categories. This is mainly due to the huge domain gap between our mostly synthetic training data and real camera test instances - 2 out of 3 are black and hence yield a larger sensor noise. Our method purely relying on depth points is not designed to cope with this issue. Consequently, our CoordinateNet fails to segment out the camera instances. To ameliorate this issue, we use 2D segmentation masks from RGB-based Mask-RCNN detection predictions provided in \cite{wang2019normalized}. When there are multiple RoIs of the camera category, we choose the one having the biggest overlap with a predicted 2D bounding box computed from our previous pose estimation. We also test our method with ground-truth segmentation masks. The results are shown in Table \ref{table:camera_seg}. Our performance significantly improves with better segmentation predictions.

\subsection{Per-Part, Per-Category Results for Articulated Object Tracking}\label{sec:per_category_articulated}

Table \ref{table:articulated_full} shows per-part, per-category articulated object pose tracking results on our synthetic dataset. In most cases, we perform better than the baseline methods and the ablated versions. We also achieve the best overall performance as shown in the main paper.

\begin{center}
\begin{table}[t]

\setlength{\tabcolsep}{3pt}
\resizebox{0.48\textwidth}{!}{

\begin{tabular}{|c|c|c|c|c|c|c|}
\hline
  Dataset & Metric & Orig. & Init.$\times$1 & Init.$\times$2 & All$\times$1 & All$\times$2 \\ \hline
\multirow{4}{*}{Rigid} & 5$^{\circ}$5cm & 62.16 & 59.64 & 55.94 & 59.83 & 58.69 \\ 
& mIoU & 64.10 & 61.40 & 57.56 & 61.30 & 60.40 \\ 
& $R_{err}$ & 5.94 & 5.95 & 5.93 & 5.81 & 5.89  \\ 
& $T_{err}$ & 7.92 & 10.23 & 10.78 & 9.82 & 13.08 \\ \hline
\multirow{4}{*}{Arti.} & 5$^{\circ}$5cm & 98.35 & 98.40 & 97.75 & 98.45 & 97.68 \\ 
& mIoU & 74.00 & 74.00 & 73.68 & 74.05 & 75.53 \\ 
& $R_{err}$ & 1.03 & 1.03 & 1.18 & 1.01 & 1.39  \\ 
& $T_{err}$ & 0.29 & 0.29 & 0.32 & 0.29 & 0.39 \\ \hline
\end{tabular}
}
\vspace{0.2mm}
\caption{\textbf{Robustness to pose errors.} Init.$\times m$ means adding $m$ times train-time errors in pose initialization, on top of the $1\times$ train-time error already used in our original setting (denoted Orig.), and All$\times m$ means adding $m$ times the errors to all estimated poses.
}
\vspace{-3mm}
\label{table:robustness}
\end{table} 
\end{center}
\vspace{-5mm}
\subsection{Robustness to Pose Errors}\label{sec:supp_robustness}
Table \ref{table:robustness} shows the performance of our model w.r.t. different amount of pose errors. We test our model under the following settings: (1) increasing the intial pose error by $1$ or $2$ times, denoted as Init.$\times 1$ and Init.$\times 2$; and (2) adding $1$ or $2$ times pose error to every previous frame's prediction, denoted as All.$\times 1$ and All.$\times 2$, to examine the robustness to pose initialization and estimation errors, respectively. For rigid objects, our performance degrades gracefully. And as shown in Fig.4 of the main paper, it is significantly more robust to noises than 6-PACK. For articulated objects, our method is very robust with less than $1$ point drop on both 5$^{\circ}$5cm and mIoU metrics. 

\subsection{Additional Visualization}
\begin{figure*}[htp]
\begin{center}
   \includegraphics[width=0.85\linewidth]{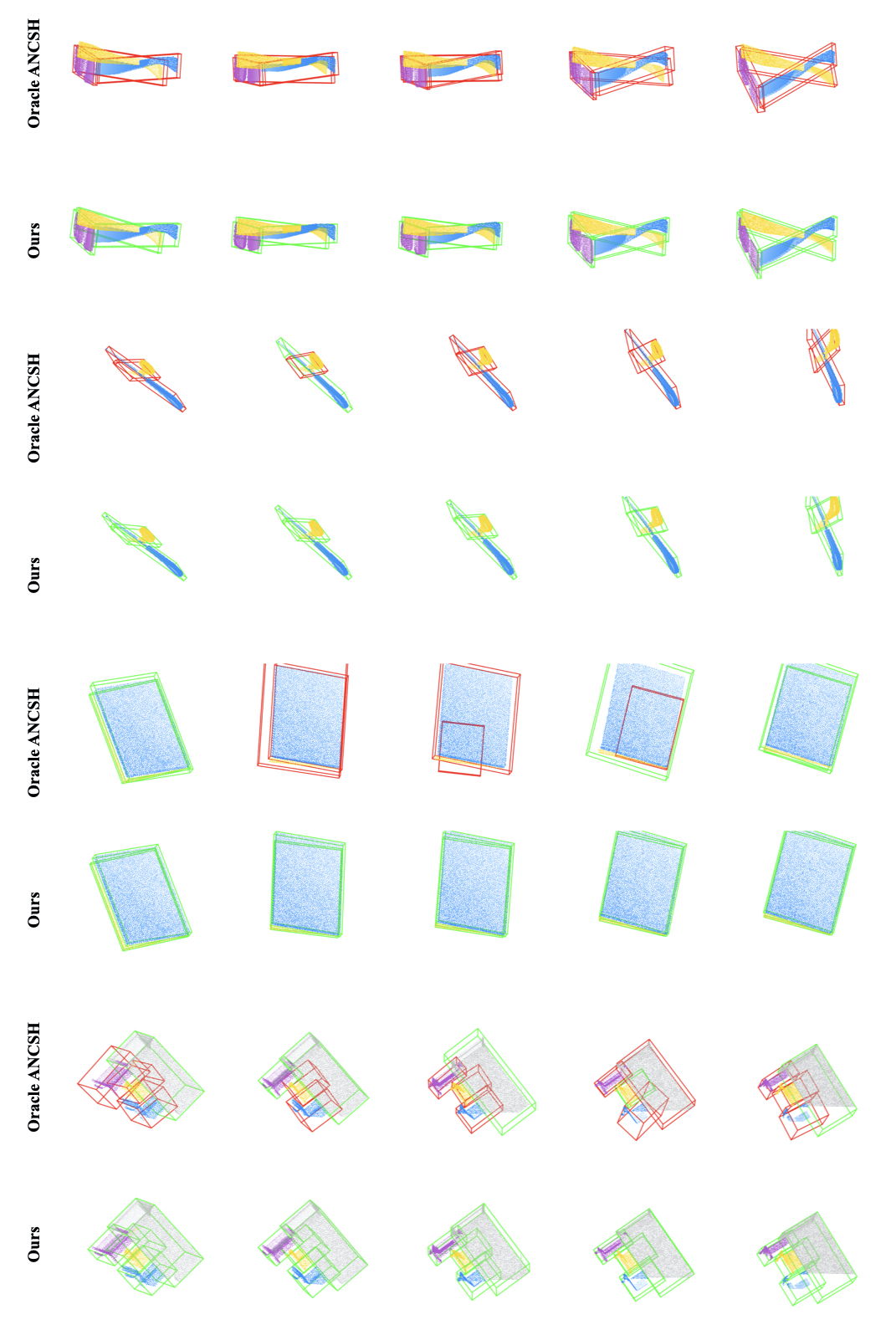}
\end{center}
   \vspace{-6mm}
   \caption{\textbf{Result visualization on the SAPIEN articulated object dataset.} Here we compare our method with oracle ANCSH, which assumes the availability of ground truth part masks.} 
   \label{fig:visualization}
\end{figure*}

\begin{figure*}[htp]
\begin{center}
   \includegraphics[width=0.85\linewidth]{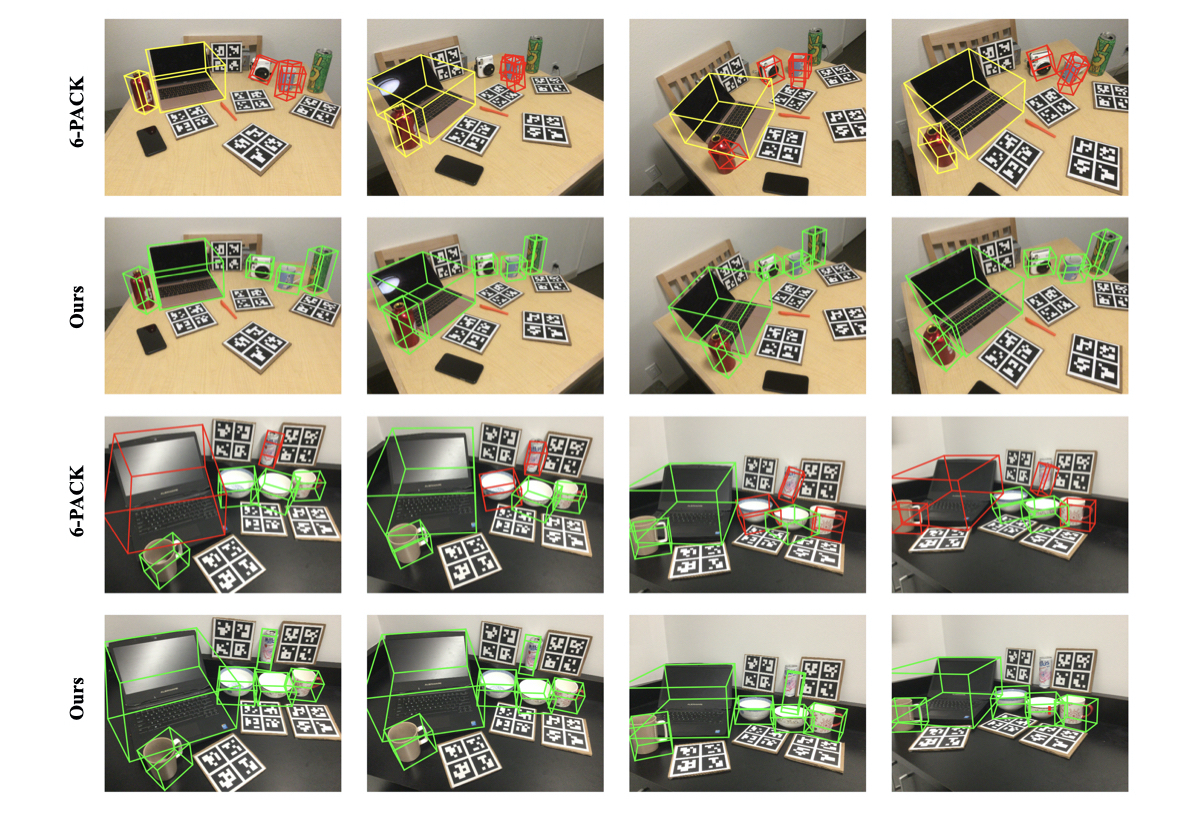}
\end{center}
    \vspace{-6mm}
   \caption{\textbf{Result visualization on the NOCS-REAL275 dataset.} Here we compare our method with 6-PACK initialized with the same pose noise as ours.} 
   \vspace{-2mm}
   \label{fig:visualization2}
\end{figure*}

\begin{figure*}[htp]
\begin{center}
   \includegraphics[width=0.85\linewidth]{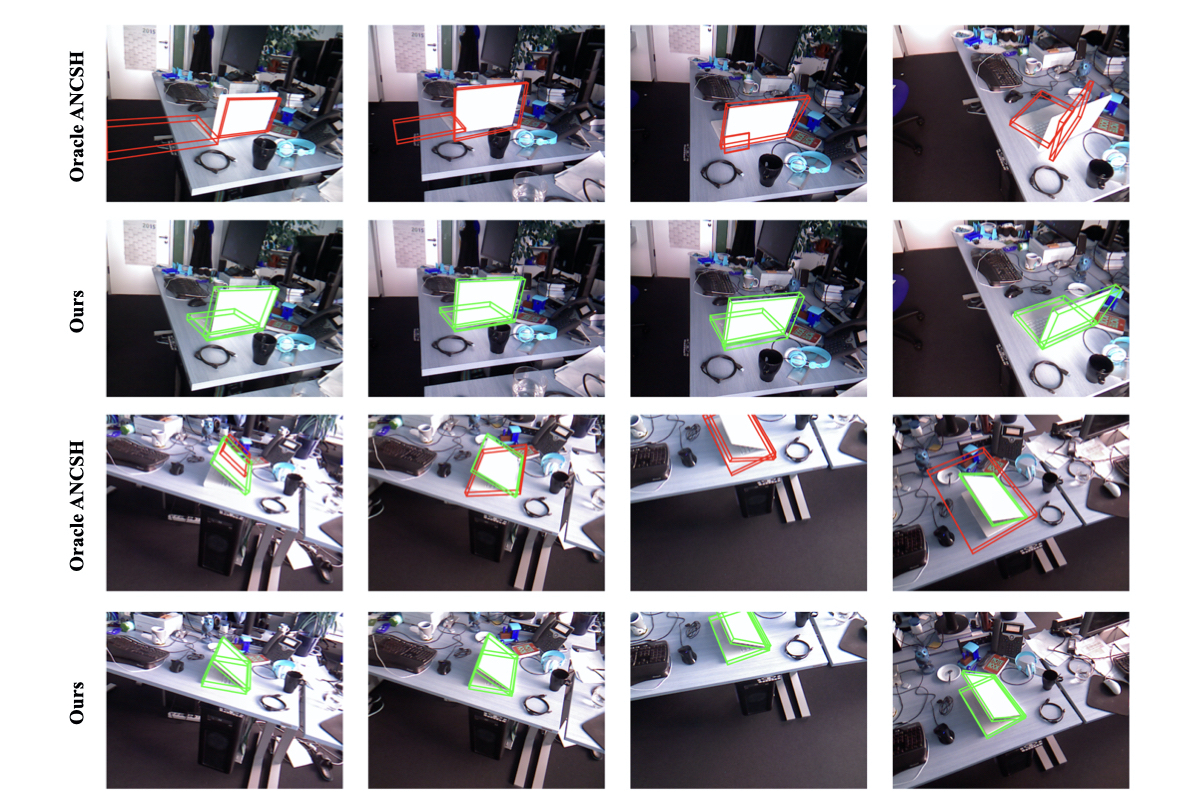}
\end{center}
\vspace{-6mm}
   \caption{\textbf{Result visualization on the real laptop trajectories from the BMVC dataset.} Here we compare our method with Oracle ANCSH under the category-level setting, where the two methods only see synthetic data from SAPIEN during training and directly test on the real data without finetuning.} 
   \label{fig:visualization3}
\end{figure*}

See Fig.\ref{fig:visualization} - \ref{fig:visualization3}  for our visual tracking results. For more visual results, please refer to our supplementary videos.

~\\
~\\

\end{appendix}

\end{document}